\documentclass{article}

\usepackage[nonatbib]{neurips_2024}

\usepackage[utf8]{inputenc} % allow utf-8 input
\usepackage[T1]{fontenc}    % use 8-bit T1 fonts
\usepackage{hyperref}       % hyperlinks
\usepackage{url}            % simple URL typesetting
\usepackage{booktabs}       % professional-quality tables
\usepackage{amsfonts}       % blackboard math symbols
\usepackage{nicefrac}       % compact symbols for 1/2, etc.
\usepackage{microtype}      % microtypography
\usepackage{xcolor}         % colors

\usepackage{tabularx}
\usepackage{multicol}
\usepackage{multirow}
\usepackage{amsmath}
\usepackage{graphicx}
\usepackage{wrapfig}
\usepackage{hwemoji}
\usepackage{listings}
\usepackage{float}
\definecolor{awesome}{rgb}{1.0, 0.13, 0.32}
\definecolor{azure(colorwheel)}{rgb}{0.0, 0.5, 1.0}
\definecolor{aureolin}{rgb}{0.99, 0.93, 0.0}
\definecolor{amber}{rgb}{0.99, 0.93, 0.0}
\definecolor{frenchrose}{rgb}{0.96, 0.29, 0.54}
\definecolor{coquelicot}{rgb}{1.0, 0.22, 0.0}
\definecolor{aliceblue}{rgb}{0.9, 0.9, 0.9}

\lstdefinelanguage{prompt}{
    frame=shadowbox,
    framerule=0.5pt,
    framesep=2pt,
    breaklines=true,
    backgroundcolor=\color{aliceblue},
    basicstyle=\fontsize{9pt}{9pt}\selectfont\ttfamily,
    commentstyle=\color{cyan},
    morecomment=[l]{//},
    moredelim=[is][\color{frenchrose}\bfseries]{<<<}{>>>},
    moredelim=[is][\color{awesome}\bfseries]{***}{***},
    moredelim=[is][\color{azure(colorwheel)}\bfseries]{///}{///},
    moredelim=[is][\color{coquelicot}\bfseries]{|||}{|||},
}

\lstdefinestyle{mystyle}{
    basicstyle=\fontsize{9pt}{9pt}\selectfont\ttfamily,
    breakatwhitespace=false,
    backgroundcolor=\color{aliceblue},
    xleftmargin=1pt,
    breaklines=true,
    breakindent=0pt,
}
\newcommand{\mycustomsymbol}{\includegraphics[width=0.8em]{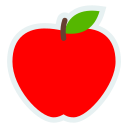}}

\title{Think out Loud: Emotion Deducing Explanation in Dialogues}

\author{
  Jiangnan Li\textsuperscript{\rm 1,2},~~~ Zheng Lin\textsuperscript{\rm 1},~~~ Lanrui Wang\textsuperscript{\rm 1},~~~ Qingyi Si\textsuperscript{\rm 1},~~~ Yanan Cao\textsuperscript{\rm 1}, \\
  \textbf{Mo Yu}\textsuperscript{\rm 2},~~~ \textbf{Peng Fu}\textsuperscript{\rm 1},~~~ \textbf{Weiping Wang}\textsuperscript{\rm 1},~~~ \textbf{Jie Zhou}\textsuperscript{\rm 2} \\
  \textsuperscript{\rm 1} Institute of Information Engineering, Chinese Academy of Sciences\\
  \textsuperscript{\rm 2} Pattern Recognition Center, WeChat AI, Tencent Inc.
}

\begin{document}

\maketitle

\begin{abstract}
Humans convey emotions through daily dialogues, making emotion understanding a crucial step of affective intelligence. To understand emotions in dialogues, machines are asked to recognize the emotion for an utterance (Emotion Recognition in Dialogues, ERD); based on the emotion, then find causal utterances for the emotion (Emotion Cause Extraction in Dialogues, ECED). The setting of the two tasks requires first ERD and then ECED, ignoring the mutual complement between emotion and cause. To fix this, some new tasks are proposed to extract them simultaneously. Although the current research on these tasks has excellent achievements, simply identifying emotion-related factors by classification modeling lacks realizing the specific thinking process of causes stimulating the emotion in an explainable way. This thinking process especially reflected in the reasoning ability of Large Language Models (LLMs) is under-explored. To this end, we propose a new task ``Emotion Deducing Explanation in Dialogues'' (EDEN). EDEN recognizes emotion and causes in an explicitly thinking way. That is, models need to generate an explanation text, which first summarizes the causes; analyzes the inner activities of the speakers triggered by the causes using common sense; then guesses the emotion accordingly. To support the study of EDEN, based on the existing resources in ECED, we construct two EDEN datasets by human effort. We further evaluate different models on EDEN and find that LLMs are more competent than conventional PLMs. Besides, EDEN can help LLMs achieve better recognition of emotions and causes, which explores a new research direction of explainable emotion understanding in dialogues. 
\end{abstract}

\section{Introduction}

\begin{wrapfigure}{r}{0.55\textwidth}
  \vspace{-0.4cm}
  \begin{center}
    \includegraphics[width=0.55\textwidth]{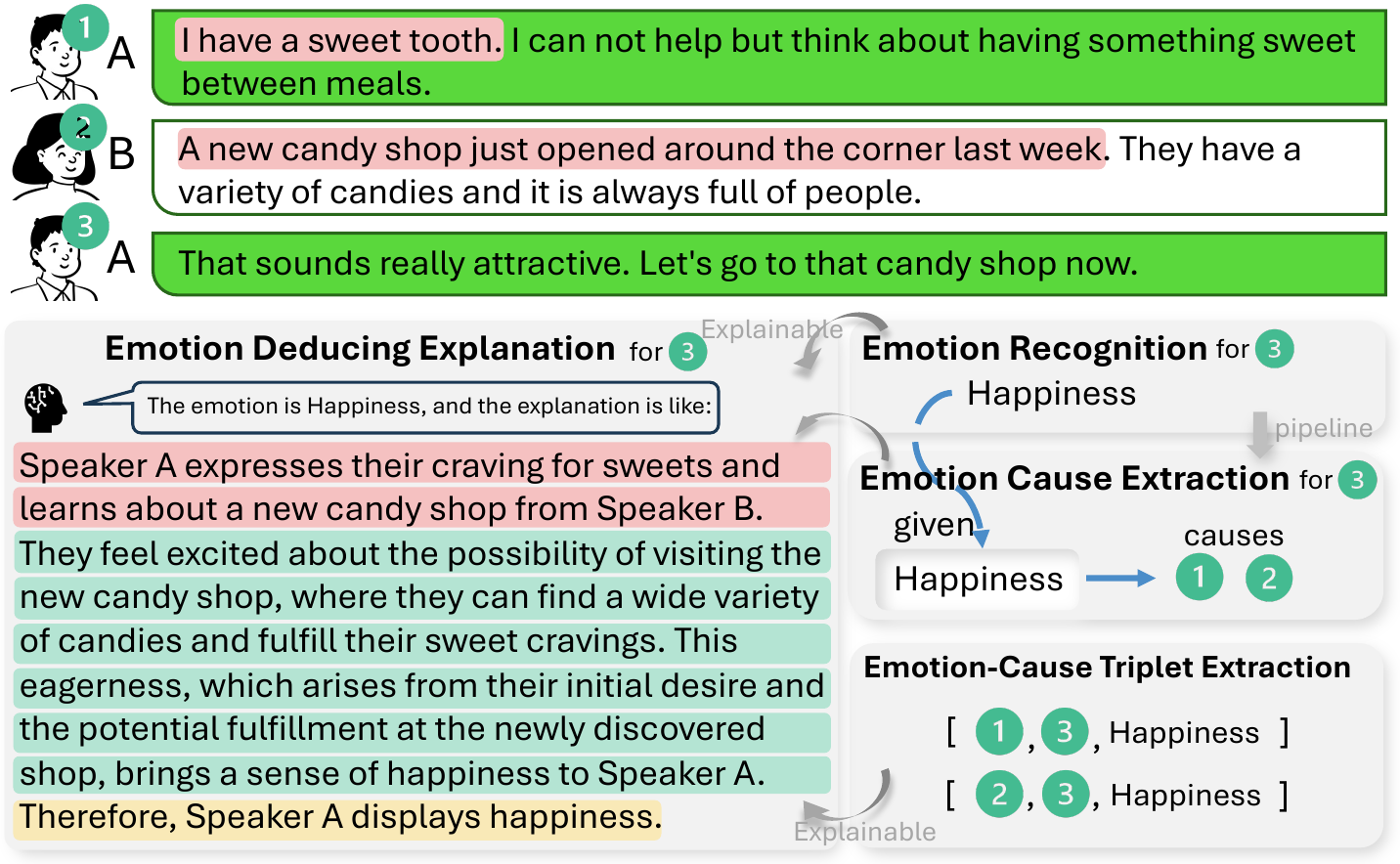}
  \end{center}
  \caption{A case of EDEN~\mycustomsymbol\ and other emotion-related classification tasks. }
  \label{start_case}
  % \vspace{-0.05cm}
\end{wrapfigure}

Understanding emotions expressed by humans in their dialogues is an inevitably essential step of intelligent affective computing for the purpose of research~\cite{ERC_Survey,ECE_Survey} and practice~\cite{depress}. Emotion and its cause are two fundamental factors of emotion understanding, boosting the emergence of Emotion Recognition in Dialogues (ERD)~\cite{IEMOCAP,DailyDialog,MELD} and Emotion Cause Extraction in Dialogues (ECED)~\cite{RECCON}. ERD predicts the emotion for an utterance, and ECED then finds causal utterances for the given emotion. However, the pipeline-style setting ignores the mutual indication of emotion and cause~\cite{ECPE}. To solve this problem, the pair extraction of emotional and causal utterances is proposed (ECPEC)~\cite{ECPEC}, but ECPEC does not predict the specific emotion of the emotional utterance. To this end, researchers~\cite{ECF,ECTEC_csk} propose to extract the triplets of these utterances and the emotion (ECTEC) as shown in Fig.~\ref{start_case}. 

Although research~\cite{KEC,KBCIN,ECTEC_csk,TFD,Bipartite} on these emotion-understanding tasks has achieved excellent achievements, classification modeling only gives outcomes of emotion-related factors, lacking in realizing how the emotion is stimulated by its causes. That is, models do not explicitly analyze the appraisal~\cite{appraisal}, which is theoretically related to human psychological activities developing from emotional triggers~\cite{aronson2005social}. This leads to models failing to deeply reason the process of emotion arousal in an explainable and explicit thinking way. Especially, with the flourishing of Large Language Models (LLMs), such an explainable thinking process is crucial for LLMs to achieve better cognition, which is under-explored. 

To achieve explainable emotion understanding, we propose a new task called ``Emotion Deducing Explanation in Dialogues'' (EDEN \mycustomsymbol). EDEN fully considers the mutual complement of emotion and cause and recognizes them simultaneously by generating an explanation text. As shown in Fig.~\ref{start_case}, only giving a dialogue whose last utterance is the target to analyze, EDEN generates the explanation text by: summarizing the emotion triggers from the dialogue context (red highlights), analyzing the speaker's inner activities towards these triggers (green highlights), and giving the specific emotion aroused from the inner activities (yellow highlights). We call this thinking process as emotion deducing. Specifically, the analysis of inner activities needs commonsense of events and mental state~\cite{SocialIQA,ATOMIC}, which sets a requirement of the reasoning ability of models. 

To support the study of EDEN, we start from the existing data resources of ECEC and ECTEC, which have provided emotion and cause annotations, and construct two datasets by human effort: DailyDialogue (EDEN-DD) and Friends (EDEN-FR). Based on the curated datasets, we adapted experiments on a range of models including conventional pretrained models (PLMs), ChatGPT, and fine-tuned LLaMA (which we propose it as EDEN-LLaMA). We show that PLMs are not competent to EDEN and the reasoning ability of LLMs can be activated by EDEN to achieve better emotion understanding. With EDEN, our method can achieve better emotion/cause recognition compared to previous models. Besides, we also show that LLMs' ability to compositional and multi-hop reasoning still requires to be strengthened. Overall, EDEN provides a new research direction of explainable emotion understanding in dialogues, which is more challenging than previous tasks.

\section{Related Work}

\textbf{Emotion Understanding in Dialogues} involves a range of classification tasks: emotion recognition (ERD)~\cite{IEMOCAP,MELD,DailyDialog}, cause (ECED)~\cite{RECCON}, emotion-cause pair (ECPEC)~\cite{ECPEC}, and triplets (ECTEC)~\cite{ECF} extractions. These tasks lack reasoning human inner activities due to the absent commonsense reasoning ability of the used PLMs~\cite{bird_cannot_fly}. To this end, a lot of works focus on introducing commonsense knowledge~\cite{ConceptNet,ATOMIC,COMET,COMET2020} to enhance this. For ERD, the knowledge is embedded in dialogue context modeling~\cite{COSMIC,SKAIG,KET}, and cooperates with topic~\cite{TODKAT} and sentimental~\cite{KI} information. For ECED, the knowledge is used to find causal clues~\cite{KEC}, is structurally modeled according to its types~\cite{KBCIN}, and is constructed to be a bipartite graph~\cite{Bipartite}. For ECTEC, the knowledge is used to enhance the understanding of implicit expressions~\cite{ECTEC_csk}. Although the introduction of knowledge can bring interpretability to some extent, the reasoning ability of models is enhanced by external knowledge supplier~\cite{COMET}, which is often noisy~\cite{KEC} and is not naturally reflected by the models themselves. With the emergence of LLMs~\cite{ChatGPT,LLaMA2}, the powerful thinking process of them can manifest such ability. Recently, ChatGPT has been used to generate commonsense knowledge for ERD~\cite{ChatGPTGenKnow}. The CoT-style explanation for emotion cause is explored~\cite{ECR_Chain}. However, there is no work providing human-curated data resources and systematic studies for explainable understanding of both emotion and cause. 

\textbf{Commonsense Reasoning and Explainability} are two factors highly related to our task. The commonsense in our task mainly focuses on events and speakers' mental state. ATOMIC~\cite{ATOMIC,COMET2020,ATOMICx10} and SocialIQA~\cite{SocialIQA} are the most popular resources for this purpose, which is usually trained for knowledge generation~\cite{COMET}, but they serve for single events without context. GLUCOSE~\cite{GLUCOSE} is proposed for contextual commonsense reasoning in stories and CICERO~\cite{CICERO,CICEROv2} is for dialogues. However, these resources only support a round of reasoning at a time, which is suitable for complicated analysis of several aspects. As for explainability, the explanation for sarcasm is proposed for single sentences~\cite{SarcasmExplanation} and dialogues~\cite{SarcasmExplanationInDialogue}. EMER~\cite{EffectGPT} explains the emotion by analyzing a video clip with a short subtitle, mainly focusing on the audio and visual modalities. These works do not explore explanations around emotion and cause in dialogues. 

\section{Our EDEN Dataset}

\subsection{Problem Definition}

% In the real scenario, dialogues move forward between speakers in real time and it is impossible to get access to future utterances. To this end, we set models to perform EDEN for every emotional utterance that is the last utterance of the real-time dialogue. 

Formally, for a dialogue $\mathcal{D}=[u_{1}, ..., u_{n}]$, the last utterance $u_{n}$ is an utterance with an unknown emotion (the target utterance). $u_{i}$ contains several factors: its turn id ${i}$, corresponding speaker $s_{i}$, and content text $t_{i}$. Now, models only know that $u_{n}$ said by $s_{n}$ shows an unknown emotion, the model should generate the explanation to deduce the emotion and guess the emotion $e_{k}$ from [`\textit{happiness} (\textit{joy})', `\textit{surprise}', `\textit{sadness}', `\textit{anger}', `\textit{disgust}', `\textit{fear}']: $Model(\mathcal{D},u_{n})\rightarrow$``The emotion is $e_{k}$. $\mathcal{E}$'', where $\mathcal{E}$ is the explanation acting as the Chain of Thought (CoT) for emotion recognition. 

\subsection{Data Collection}

% As there is no available emotion deducing explanation in dialogues, we intend to construct these kinds of resources by further human annotation on existing datasets. The existing dataset should satisfy two conditions: First, the target utterance should be labeled with a specific emotion; Second, the emotion-causal utterances should also be annotated beforehand so that further human annotations can avoid the burden of locating emotion causes in the dialogue. With these two kinds of pre-annotated labels, humans only need to focus on explaining how the emotion triggers induct the emotion. 

As there is no available data for EDEN, we intend to construct the resource by further human annotation on existing datasets. The existing dataset should satisfy two conditions: First, utterances should be labeled with their emotions; Second, causal utterances should also be annotated beforehand. 

For this purpose, we collect two datasets: RECCON~\cite{RECCON} in emotion cause extraction and ECF~\cite{ECF} in emotion-cause triplet extraction. From RECCON, we pick up its DailyDialogue part, which consists of considerable dyadic dialogues, denoted as ``DD''. As for ECF, we directly use its whole unimodal data. As the data of ECF are multi-party dialogues from a popular TV series called ``\textit{Friends}'', we denote it as ``FR''. 

% We collect two open-source datasets: RECCON~\cite{RECCON} in emotion cause extraction (ECED) and ECF~\cite{ECF} in emotion-cause triplet extraction (ECTEC). From RECCON, we pick up its DailyDialogue part, which consists of considerable dyadic dialogues, and we denote it as ``DD''. As for ECPEC, we directly use its whole unimodal data. As the data of ECPEC are multi-party dialogues from a popular TV series called ``\textit{Friends}'', we denote it as ``FR''. 

% \textbf{Data Preparing.}~~~~Researchers studying these two datasets pay more attention to constructing graphs upon utterances and predicting the emotion label or emotion causal utterances for all target utterances in a dialogue at once, which is definitely unable to fit into our setting. For our setting, only one target utterance along with its history is analyzed and produces only a piece of explanation each time. Therefore, we construct $m$ samples if a dialogue contains $m$ non-neutral target utterances. 

\textbf{Data Preparing.}~~~~In our setting, only the target utterance along with its history are input for analysis. Therefore, we construct $m$ samples if a dialogue contains $m$ non-neutral utterances. Furthermore, we find that most of the causal utterances are located near the target utterance and less than 15 turns away. To this end, we set the history context window to 15. Those causal utterances out of the window will be truncated, and if a target utterance does not have in-window causal utterances, the sample will be removed. We prepared 5331 samples for DD where the number of training/dev/test data is 4645/157/529, and 6741 samples for FR, whose case is 4874/591/1276. 

% Furthermore, we find that almost all emotion-causal utterances are located near the target utterance and more than 99\% of these causal utterances are less than 15 turns away. To this end, we set the history context window to 15 (including the target utterance itself). Those causal utterances out of the window will be truncated, and if a target utterance does not have in-window causal utterances, the sample will be removed because these samples are rare and not our research attention. After the sample construction and filtering, we prepared 5331 samples for DD where the number of training/dev/test data is 4645/157/529, and 6741 samples for FR where the number of training/dev/test data is 4874/591/1276. 

For human annotation, the burdens of writing the explanation from scratch are heavy and exhausting. Therefore, we utilize ChatGPT to help with the annotation. We input a sample with the emotion and its causal utterances to ChatGPT and instruct ChatGPT to generate the original explanation of how causes lead to the emotion in a Chain-of-Thought style. Following \cite{summary_LLMs}, we also use a two-round prompt to let ChatGPT first elaborate on the analysis and then summarize the analysis. The prompt is shown in Fig.~\ref{two_round}, Appendix~\ref{appendix_prompt}. To further assist human annotators, we also utilize ChatGPT to generate the topic of the dialogue, which can help annotators quickly understand the dialogue. 

% For human annotation, if we solely provide the emotion and its causal utterances, the burdens of annotations will be heavy and easily lead to annotators' tiredness. Instead of writing the explanation for all samples, we utilize ChatGPT to help with the annotation, i.e., ChatGPT generates the original explanation. However, as shown by (cite), LLMs like ChatGPT are not experts in predicting a speaker's emotion or finding the exact emotion causes in the dialogue. Hiring ChatGPT to generate the explanation directly will produce low-quality data. To this end, we input the emotion and its causal utterances to ChatGPT and instruct it to only generate the analysis of how causes lead to the emotion in a Chain-of-Thought style. Following (), we also use a two-round prompt to let ChatGPT first elaborate on the analysis and then summarize the analysis. The prompt is shown in Fig.~. To further assist human annotators, we also utilize ChatGPT to generate the topic of the dialogue, which can help annotators quickly understand what content the dialogue is based on. 

% from the given causal utterances to the specific emotion

\textbf{Human Annotations.}~~~~As we have mentioned in previous paragraphs, ChatGPT will generate the original analysis. To this end, annotators need to read the analysis and assess its reasonability. If the analysis makes sense by providing a reliable reasoning chain, annotators can save the labor by only paraphrasing the original analysis to the format of our task's requirement. Otherwise, annotators rewrite the explanation according to their own reasoning. 

We hired a group of Chinese undergraduates who are in major of computer sciences to make sure they can easily understand the study of computational linguistics and AI terminologies. Besides, all the annotators have passed the CET-6 test\footnote{College English Test-6. A national standard test that certifies the examinee has reached the English level of non-English major postgraduates in China, which sets a high requirement for undergraduates. } at least to ensure their English thinking and writing skills. The annotators are trained by the following simplified guidelines:

(1) The emotion triggers are required to be extracted from the given causal utterances based on the dialogue. These triggers can be events happening in the dialogue~\cite{RECCON}, and even the speaker's external verbal words when the dialogue provides less information for an exact event. For these events, the annotators are encouraged to refine and summarize them instead of directly copying spans from the dialogue (copying spans is still acceptable). 

(2) The analysis from the emotion trigger to the target emotion should focus on the speaker's psychological activities towards the events, requiring humans' social commonsense about the \textit{effect} of the events~\cite{ATOMIC}; speakers' \textit{intent}, \textit{reaction}, \textit{attribute}, and \textit{behavior}~\cite{ATOMIC}.

(3) Summarize emotion triggers, organize the analysis, and give the emotion, as illustrated in Fig.~\ref{start_case}. 

\begin{wraptable}{r}{5.5cm}
\vspace{-0.5cm}
\centering
\caption{The average scores for 3 aspects to verify the data quality. }
\vspace{0.2cm}
\scalebox{0.6}{
\begin{tabular}{l|c|c|c}
\toprule
                     & \textit{fluency} & \textit{correctness} & \textit{rationality} \\ \midrule
DD                   & 2.960            & 2.927                & 2.927                \\
FR                   & 3.000            & 2.907                & 2.913                \\ \midrule
Fleiss' Kappa for DD & 0.923            & 0.823                & 0.806                \\
Fleiss' Kappa for FR & 1.000            & 0.701                & 0.791                \\ \bottomrule
\end{tabular}
}
\label{quality}
\end{wraptable}

Before the formal annotation, annotators are assigned a subset of samples separated into 5 parts for the training. They annotate one part and submit it to a conversational AI expert. This part is passed when the expert approves or otherwise is returned for revision. After an average of 3 iterations, the annotator can reach a standard. For the formal annotations, annotators rate ChatGPT's analysis and produce the final explanation. The quality of ChatGPT's analysis is ranked by 3 levels: good (3 scores), okay (2 scores), and needing improvement (1 score). The annotating interface is shown in Fig.~\ref{screenshot}, Appendix~\ref{appendix_interface}. 

% are randomly assigned samples, 

As errors are always unavoidable, we utilize ChatGPT to correct grammar errors and typos. More importantly, ChatGPT is asked not to change the original structure and meaning. After the post-process, we evaluated a subset and found that ChatGPT can remove these errors with minor changes. 

\begin{figure*}
    \centering
    \scalebox{0.23}{
    \includegraphics{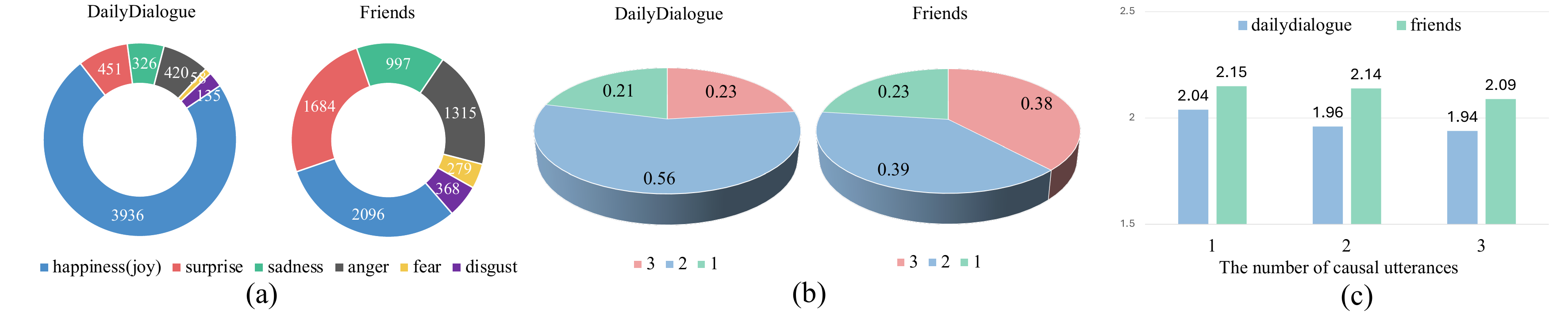}
    }
    \caption{(a) the emotion distributions in DD and FR; (b) The number of samples whose original ChatGPT analysis is scored to 1 score to 3 scores; (c) The average scores of ChatGPT analysis in the samples with different numbers of causal utterances. }
    \label{more_statistics}
    \vspace{-0.4cm}
\end{figure*}

\textbf{Annotating Quality.}~~~~To evaluate the quality of the annotated data, we randomly and respectively pick up 50 samples from DD and FR. We hire 3 conversational AI researchers as inspectors. The verification mainly focuses on 3 aspects: \textit{fluency} - whether the explanation is fluent and has no clear description; \textit{correctness} - whether the emotion triggers are extracted from the causal utterances based on the dialogue; \textit{rationality} - whether the explanation provides a reasonable analysis from the triggers to the target emotion. The three aspects can be scored: [1, 2, 3]. The average scores of the 3 aspects and the agreement between inspectors are illustrated in Tab.~\ref{quality}. The results show that the annotated data achieves substantial or almost perfect agreements among inspectors. 

% and the quality is satisfying

\begin{wraptable}{r}{5.2cm}
\vspace{-0.5cm}
\centering
\caption{The statistics of datasets.}
\vspace{0.2cm}
\scalebox{0.7}{
\begin{tabular}{l|cc|c}
\toprule
EDEN           & DD    & FR    & Total \\ \hline
num. samples   & 5331  & 6741  & 12072 \\ \hline
avg. dialogue len. & 6.26  & 7.28  & 6.83  \\ \hline
avg. utterance len. & 15.63 & 12.54 & 13.79 \\ \hline
avg. explanation len. & 76.16 & 68.34 & 71.79 \\ \bottomrule
\end{tabular}
}
\label{statistics}
\end{wraptable}

\textbf{Data Statistics.}~~~~The basic data statistics are shown in Tab.~\ref{statistics}. It can be seen that the real-time dialogue length of FR is longer than that of DD. However, the utterances in DD are more informative than FR as utterances in FR often contain meaningless exclamation words like ``ah'' and ``yeah'', informal expressions, and American slang and memes because the data is from American comedy. For more illustrations, as shown in Fig.~\ref{more_statistics} (a), the emotion distributions are different in DD and FR, where DD shows a more evident long-tail distribution than FR. 

% the positive emotion ``happiness'' is dominant in 
% and the emotion distribution in FR is more balanced

As for annotators' evaluation to ChatGPT's original analysis, we compute the average scores for DD and FR separately. The score for DD is 2.03 and for FR is 2.14, where the detailed score distribution can be viewed in Fig.~\ref{more_statistics} (b). The reason for FR having more good ChatGPT analyses is that dialogues in FR provide less information. Story understanding in \textit{Friends} requires more context in the whole scene shot or even global information~\cite{TVSHOWGUESS}. However, samples in FR are short clips, and annotators tend to trust ChatGPT's analysis when few clues can be found in dialogues. In general, the result shows that ChatGPT possesses an acceptable ability to analyze inner activities and reason commonsense, but it still has space for improvement. 

% by only textual modal 

% compared to human's rationality score of 2.92 (Tab.~\ref{quality})

Besides, we also compute the average scores of samples with different numbers of causal utterances, whose results are depicted in Fig.~\ref{more_statistics} (c). The reason for only considering up to 3 causal utterances is that the number of samples with >3 causal utterances is far fewer. As seen from the figure, ChatGPT tends to perform slightly worse with the number of causal utterances increasing.

% , which is natural because increasing factors can make the analysis complicated. 

% Besides, we limit the length of the explanation, which can lead to more refined deducing. 

\begin{table}[]
\centering
\caption{Comparison between EDEN and other affection-related dataset. $*$ denotes that only partial data has such a function. $\dagger$ denotes that commonsense reasoning does not focus on how causes lead to emotion. ``t'' denotes ``textual'' modality; ``a'' for ``audio''; ``v'' for ``visual''. }
\scalebox{0.6}{
\begin{tabular}{l|cccccccc}
\toprule
dataset       & task format                  & modality & dialogue     & \begin{tabular}[c]{@{}c@{}}emotion\\ predict\end{tabular}  & \begin{tabular}[c]{@{}c@{}}emotional\\ cause realize\end{tabular} & \begin{tabular}[c]{@{}c@{}}commonsense\\ reason\end{tabular} & explanation  & CoT style    \\ \hline
IEMOCAP~\cite{IEMOCAP}       & emotion classification       & t,a,v    & $\checkmark$ & $\checkmark$     &                         &                    &              &              \\
% MELD~\cite{MELD}          & emotion classification       & t,a,v    & $\checkmark$ & $\checkmark$     &                         &                    &              &              \\
DailyDialogue\cite{DailyDialog} & emotion classification       & t        & $\checkmark$ & $\checkmark$     &                         &                    &              &              \\ \hline
RECCON~\cite{RECCON}      & cause classification         & t        & $\checkmark$ &                  & $\checkmark$            &                    &              &              \\
ConvECPE~\cite{ECPEC}      & cause classification (pair)  & t        & $\checkmark$ &                  & $\checkmark$            &                    &              &              \\ 
ECF~\cite{ECF}           & cause classification (triplets) & t,a,v    & $\checkmark$ &  $\checkmark$       & $\checkmark$            &                    &              &              \\ \hline
CICEROv2~\cite{CICEROv2}        & commonsense reason           & t        & $\checkmark$ & $\checkmark^{*}$ & $\checkmark^{*}$        & $\checkmark^{\dagger}$   &              &              \\
SED~\cite{SarcasmExplanationInDialogue}          & sarcasm explanation          & t,a,v    & $\checkmark$ &                  &                         &                    & $\checkmark$ &              \\
EMER~\cite{EffectGPT}          & emotion explanation          & t,a,v    &              & $\checkmark$     &                         &                    & $\checkmark$ & $\checkmark$ \\ \hline
EDEN \mycustomsymbol (ours)    & emotion deducing explanation & t        & $\checkmark$ & $\checkmark$     & $\checkmark$            & $\checkmark$       & $\checkmark$ & $\checkmark$ \\ \bottomrule
\end{tabular}
}
\label{task_comparison}
\vspace{-0.3cm}
\end{table}

\subsection{Comparisons with Other Tasks }

Tab.~\ref{task_comparison} compares EDEN and other highly related datasets. As shown in the table, EDEN predicts both emotion and cause, which can be regarded as a substitution for these classification tasks. Although CICERO provides two types of commonsense for emotion and cause independently, emotion reasoning still directly predicts emotion, and cause reasoning is not exclusive to emotion. Besides, CICERO reasons a type of commonsense at a time, which cannot make more challenging reasoning from emotional causes to emotion as EDEN does. EMER also explains emotions, but it is for a single sentence, not suitable for dialogues. The explanation of EMER mainly focuses on audio and visual modalities, and the analysis of text does not consider emotion causes or specific inner activities of speakers. Conversely, EDEN is proposed to deeply model cause and emotion through mental state analysis and other types of commonsense reasoning. 

\section{Method}

\subsection{Conventional Models}

\textbf{Small SOTA methods in ERD and ECED.}~~~~As EDEN predicts emotion and cause together, the performance of SOTA methods to identify these two factors can be a reference. For ERD, TFD~\cite{TFD}, a framework to reduce major label biases, and BHG~\cite{Bipartite}, which constructs a bipartite graph for utterances and commonsense knowledge, are selected. For ECED, BHG is still the SOTA method. 

\textbf{Pretrained Language Models.}~~~~Exploring whether previous pretrained models are competent to EDEN is crucial. BART~\cite{BART} and GPT2~\cite{GPT2} can be representative for this purpose. For BART, we form the source and target as: ``[BOS] $\mathcal{C}_{\leq t}$ [SEP] $s_i$: $t_i$ [EOS] $\rightarrow$ [BOS] $\mathcal{A}$ [EOS]'', where $\mathcal{C}_{\leq t}$ is the dialogue; $t_i$ is the target utterance and $s_i$ is the target speaker; $\mathcal{A}$ is the output containing the explanation. For GPT2, we form the source and target as: ``Dialogue is: $\mathcal{C}_{\leq t}$; The target utterance is: $s_i$: $t_i$. Your explanation to $s_i$'s emotion at the target snippet: $\rightarrow$\ $\mathcal{A}$ <$\mid$endoftext$\mid$>''. 

\subsection{ChatGPT with In-Context Learning}

As the most famous LLM, ChatGPT demonstrates excellent reasoning ability to text understanding. We study the performance of ChatGPT. To achieve this, we have tried a zero-shot style template to prompt it, but the outcome shows extremely low scores on some automatic metrics. Therefore, we mainly focus on In-Context Learning~\cite{GPT3} and denote the method as ChatGPT $N$-shots. Instead of randomly sampling demonstrations from the training set, we utilize Sentence-BERT~\cite{SBERT} to encode the embedding of the target utterance $t_i$ of every sample; we further compute the similarity scores between them and rank training samples accordingly; the top-N samples are selected as demonstrations. The prompt template for ChatGPT 5-shots is illustrated in Fig.~\ref{chatgpt_prompt}, Appendix~\ref{appendix_prompt}. 

\subsection{EDEN-LLaMA}

Besides ChatGPT, we also explore the reasoning ability of LLaMA2~\cite{LLaMA2}. As EDEN provides training data, we fine-tune LLaMA2 with LoRA adaption~\cite{LoRA}. For LoRA fine-tuning, we set the instruction as Fig.~\ref{eden_llama_prompt}, Appendix~\ref{appendix_prompt}. In the instruction, several factors, that may affect the behavior of LLaMA2, will be considered: the candidate emotion list, the target speaker, and the identification of the target utterance. We denote the fine-tuned LLaMA2 as EDEN-LLaMA. 

\textbf{Hybrid training.}~~~~Although we evaluate EDEN-DD and EDEN-FR independently, the emotion labels of them share the Ekman's emotion theory~\cite{ekman}. We train EDEN-LLaMA using all the training data of DD and FR, and denote it EDEN-LLaMA hybrid. It will not directly compare with others.

\textbf{Supplementary commonsense pretraining.}~~~~EDEN requires the ability of commonsense reasoning to analyze a speaker's inner activities. Although LLMs have such an ability, we are still curious about whether additional training on commonsense knowledge bases can further activate such an ability. To this end, we collect three types of commonsense resources focusing on events: ATOMIC~\cite{ATOMIC}, GLUCOSE~\cite{GLUCOSE}, CICEROv2~\cite{CICEROv2}. ATOMIC focuses on events in a single sentence; GLUCOSE executes contextual reasoning in a story context; CICEROv2 is in a dialogue context. To perform instruct tuning, we construct a template pool for them, and every piece of knowledge can transfer into an instruction. We adapt instruct pretraining on them, and the trained LoRA continues EDEN training. We denote the method as EDEN-LLaMA sup. 

% for training by randomly sampling a template

\section{Experiments}

\subsection{Evaluating Metrics}

\textbf{Automatic metrics for generation.}~~~~We conventionally use BLEU, ROUGE-L, METEOR, and CIDEr scores like other generation tasks. Specifically, the speaker name in EDEN-DD is formed like ``Speaker A'' and we think this may lead to a relatively high BLEU-($\leq$2) score. Therefore, BLEU-(3/4) is used for EDEN-DD. EDEN-FR does not have such a case, and BLEU-(2/3) is used. 

\textbf{Metrics for emotion and cause recognition.}~~~~The emotion is easy to extract. However, the problem is that ChatGPT sometimes does not follow instructions to pick up the emotion from the emotion list. To reduce this, we train an emotion evaluator (based on RoBERTa~\cite{RoBERTa}) on EDEN to classify an explanation, whose emotion is not in the emotion list. The intent is that the explanation will provide clear emotion-related expressions for the evaluator to learn and use. Emotion is evaluated by a weighted F1 score denoted as EF. As for cause, the explanation does not copy original causal utterances but summarizes them, which makes the direct evaluation impossible. Nevertheless, the summarized causes are still semantically similar to causal utterances. Therefore, we train a cause evaluator to predict 0/1 with the input of ``$\mathrm{[CLS]}\ \ u_{t} \ \mathrm{[SEP]}\ u_{c} \ \mathrm{[SEP]}\ \mathcal{C}_{\leq t} \ \mathrm{[SEP]}\ \mathcal{E}\ \ \mathrm{[SEP]}$'', where $u_{t}$ is the target utterance; $u_{c}$ is a candidate utterance; $\mathcal{E}$ is the explanation. To this end, the evaluator will learn the correlation if a causal utterance is mentioned in the explanation. Based on the outcome of the Cause Evaluator, the cause is evaluated by an F1 score denoted as CF. 

\textbf{Reasonableness.}~~~~We define reasonableness as how reasonable and correct the explanation deduces the emotion from causes, which comprehensively considers the correctness of emotion prediction, cause extraction, and analyzing process. It ought to be done by humans, requiring huge labor. Instead, we follow G-EVAL~\cite{GEVAL} to evaluate it using GPT4, which is said to have good human alignment. We illustrate reasonableness (1$\leq\cdot\leq$3) in Fig.~\ref{reasonableness_prompt}, Appendix~\ref{appendix_prompt}, which provides the detailed criteria. 

\subsection{Implementations}

We use BART-large, GPT2-large, LLaMA2-7b-chat, LLaMA2-13b-chat for EDEN. For the emotion evaluator, we use RoBERTa-large, which will train 5 times for average; for the cause evaluator, due to the sequence length, we use Longformer-large~\cite{RoBERTa}, which also trains 5 times for average. As for Reasonableness, on the 100 randomly sampled samples, GEVAL will be done 20 times for average. More details of implementation can be seen in Appendix~\ref{implement}. 

\subsection{Experiment Outcomes}

\begin{table}[]
\centering
\caption{Results on EDEN-DD. B-3/4 denotes BLEU-3/4; R-L denotes ROGUE-L; MT denotes METEOR; Cr is CIDEr; EF is weighted Emotion F1; CF is Cause F1; Rt denotes Reasonableness.}
\scalebox{0.71}{
\begin{tabular}{l|c|c|c|c|c|c|c|c}
\toprule
Method                   & B-3    & B-4    & R-L    & MT     & Cr     & EF     & CF     & Rb \\ \midrule\midrule
\multicolumn{9}{c}{\small \textit{Small SOTA methods in ERD and ECED}} \\
TFD                   & -      & -      & -      & -      & -      & 84.86 & -      &  -    \\
BHG                   & -      & -      & -      & -      & -      & 87.37 & \textbf{69.40} &  -    \\
LLaMA-7b-EmoGuess     & -      & -      & -      & -      & -      & 87.77 & -      &       \\
CauseEvaluator (Gold) & -      & -      & -      & -      & -      & -      & 75.31 &  -    \\
\midrule
\multicolumn{9}{c}{\small \textit{Fine-tuned PLM}} \\
BART-large            & 14.93 & 10.37 & 30.53 & 35.46 & \underline{20.91} & 87.18 & 66.16 & 2.09 \\
GPT2-large            & 13.53 & 9.18 & 29.59 & 33.24 & 14.94 & 89.19 & 64.90 & 2.01 \\ \midrule
\multicolumn{9}{c}{\small \textit{In-context Learning LLMs}} \\
% ChatGPT 0-shot        & 0.0660 & 0.0389 & 0.2094 & 0.3365 & 0.0048 & 0.7892 & 0.6306 &      \\
ChatGPT 1-shot        & 10.60 & 6.49 & 26.16 & 37.01 & 4.80 & 77.16 & 66.33 & 2.06 \\
ChatGPT 5-shots       & 12.39 & 7.98 & 28.20 & \textbf{39.58} & 6.22 & 81.22 & 66.59 & 2.26 \\ \midrule
\multicolumn{9}{c}{\small \textit{Fine-tuned LLMs}} \\
EDEN-LLaMA-7b         & \textbf{19.71} & \textbf{13.62} & \underline{33.72} & \underline{39.01} & 19.69 & 89.38 & 68.29 & \textbf{2.53} \\
EDEN-LLaMA-13b        & \underline{19.54} & \underline{13.48} & 33.32 & 38.96 & 20.41 & \textbf{90.17} & 68.48 & \textbf{2.53} \\
EDEN-LLaMA-7b sup     & 19.43 & 13.43 & \textbf{34.10} & 38.68 & \textbf{23.13} & 89.36 & \underline{68.95} & 2.49 \\
EDEN-LLaMA-13b sup    & 19.37 & 13.24 & 33.31 & 38.54 & 20.74 & \underline{89.76} & 68.54 & \underline{2.52} \\
\hline
EDEN-LLaMA-7b hybrid  & 19.92 & 13.89 & 34.15 & 39.15 & 22.00 & 90.40 & 69.71 & 2.46 \\
EDEN-LLaMA-13b hybrid & 19.87 & 13.84 & 33.96 & 38.86 & 21.14 & 91.66 & 70.11 & 2.53 \\ 
% Mix-LLaMA-7b          & 0.2033 & 0.1426 & 0.3424 & 0.3915 & 0.2033 & 0.8857 & 0.6923 & 2.38 \\ 
\bottomrule
\end{tabular}
}
\label{tab_eden_dd_1}
\vspace{-0.5cm}
\end{table}

% To further show the improvement-required reasoning ability of PLMs

\begin{figure}
    \centering
    \includegraphics[width=1.0\textwidth]{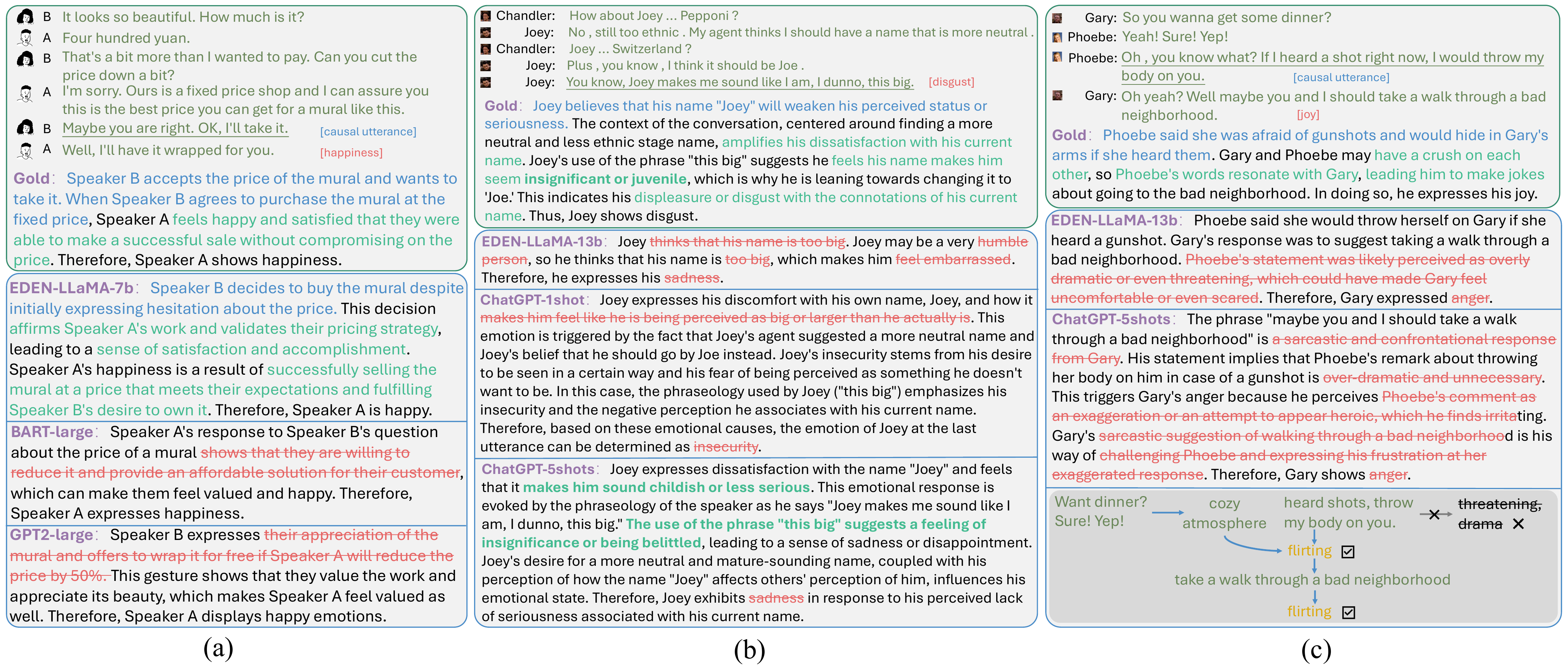}
    \vspace{-0.5cm}
    \caption{(a) A case of BART and GPT2 reasoning with factual errors. (b) A case of the activation of ChatGPT's reasoning ability by ICL. (c) A case to show that LLaMA and ChatGPT make shallow analysis. In figures, ``{\color{red} text}'' denotes the wrong analysis.}
    \label{eden_cases}
    \vspace{-0.5cm}
\end{figure}

\textbf{Main Results.}~~~~The performance of different methods on EDEN-DD and EDEN-FR is shown in Tab.~\ref{tab_eden_dd_1} and Tab.~\ref{tab_eden_fr_1}. First, let us pay attention to BART and GPT2. Their results of automatic metrics on both datasets show no advantage to EDEN-LLaMA, and the number of trained parameters of these PLMs is way larger than that of EDEN-LLaMA, which indicates that PLMs are not capable of EDEN. The reason may be that PLMs lack powerful reasoning ability and depend on co-occurrence features between emotion and some emotional expressions. This can be seen from their good performance in emotion recognition while low reasonableness scores. Accordingly, we illustrate a case that PLMs can correctly predict the emotion but make factual errors in reasoning in Fig.~\ref{eden_cases} (a). In this case, BART and GPT-2 wrongly analyze that A reduced the price for B. Conversely, EDEN-LLaMA correctly explains A's inner satisfaction with their pricing strategy. 

\begin{wraptable}{r}{10.2cm}
\vspace{-0.4cm}
\centering
\caption{Results on EDEN-FR.}
\vspace{0.3cm}
\scalebox{0.71}{
\begin{tabular}{l|c|c|c|c|c|c|c|c}
\toprule
Method                   & B-2    & B-3    & R-L    & MT     & Cr     & EF     & CF     & Rb \\ \midrule\midrule
\multicolumn{9}{c}{\small \textit{Small SOTA methods in ERD and ECED}} \\
TFD                   & -      & -      & -      & -      & -      & 53.64 & -      &  -    \\
BHG                   & -      & -      & -      & -      & -      & 64.83 & 74.88 &  -    \\
LLaMA-7b-EmoGuess     & -      & -      & -      & -      & -      & 65.54 & -      &  -    \\
CauseEvaluator (Gold) & -      & -      & -      & -      & -      & -      & 82.93 &  -    \\
\midrule
\multicolumn{9}{c}{\small \textit{Fine-tuned PLM}} \\
BART-large            & 18.17 & 11.41 & 27.20 & 30.35 & 23.18 & 64.56 & 73.72 & 1.58 \\
GPT2-large            & 16.22 & 9.98 & 25.17 & 27.79 & 17.84 & 66.05 & 72.36 & 1.41 \\ \midrule
\multicolumn{9}{c}{\small \textit{In-Context Learning LLMs}} \\
% ChatGPT emotion first & 0.0693 & 0.0365 & 0.1654 & 0.2786 & 0.0001 &    &    &    \\
% ChatGPT 0-shot        & 0.0693 & 0.0365 & 0.1654 & 0.2786 & 0.0001 & 0.5974 & 0.7065 &    \\
ChatGPT 1-shot        & 12.26 & 6.79 & 21.85 & 31.54 & 3.12 & 57.37 & 73.27 & 1.67 \\
ChatGPT 5-shots       & 14.95 & 8.79 & 24.45 & \textbf{34.20} & 7.01 & 60.39 & 74.84 & \underline{1.87} \\ \midrule
\multicolumn{9}{c}{\small \textit{Fine-tuned LLMs}} \\
EDEN-LLaMA-7b         & 21.01 & 13.57 & 29.35 & 31.91 & 27.75 & 69.40 & 75.88 & \textbf{1.89} \\
EDEN-LLaMA-13b        & \underline{21.29} & \underline{13.89} & \underline{29.58} & 32.27 & 27.12 & \underline{71.15} & \underline{76.02} & 1.85 \\
EDEN-LLaMA-7b sup     & 21.17 & 13.72 & 29.48 & 32.25 & \underline{28.04} & 70.68 & 75.75 & 1.84 \\
EDEN-LLaMA-13b sup     & \textbf{21.36}  & \textbf{14.00} & \textbf{29.71} & \underline{32.33} & \textbf{28.43} & \textbf{72.71} & \textbf{76.74} & 1.86 \\
\hline
EDEN-LLaMA-7b hybrid  & 21.09 & 13.49 & 29.03 & 31.95 & 26.27 & 69.42 & 74.94 & 1.82 \\
EDEN-LLaMA-13b hybrid & 21.72 & 14.06 & 29.61 & 32.64 & 26.91 & 71.62 & 76.05 & 1.94 \\
% Mix-LLaMA-7b          & 21.37 & 13.73 & 28.50 & 32.10 & 24.95 & 70.30 & 75.51 & 1.80 \\
\bottomrule
\end{tabular}
}
\label{tab_eden_fr_1}
\vspace{-0.2cm}
\end{wraptable}

Second, we analyze the performance of ChatGPT. Although ChatGPT is not fine-tuned for EDEN, ChatGPT shows acceptable results in automatic metrics. However, its EF is not as good as other methods because ChatGPT sometimes does not follow the instruction to select an emotion from the emotion list. The reason may be that ChatGPT has learned massive data with different emotion systems. For example, it predicts ``\textit{curiosity}'', ``\textit{disappointment}'', which may come from GoEmotions~\cite{GoEmotions}. These emotions are easily confused with ``surprise'', ``happiness'', and other negative emotions. As for the CF, ChatGPT tends to analyze history utterances with emotions while a history utterance with an emotion is not necessarily to be a cause. This makes ChatGPT have a lower precision score for the cause compared with EDEN-LLaMA. As for ChatGPT's reasoning ability, with the number of demonstrations increasing, all the metrics increase, especially the score of reasonableness. This indicates that more demonstrations in in-context learning can activate ChatGPT's reasoning ability. To illustrate this, Fig.~\ref{eden_cases} (b) gives a case of analyzing ``Joe'' and ``Joey''. EDEN-LLaMA and ChatGPT 1-shot wrongly explain that ``Joey'' is a big name. However, ChatGPT 5-shots can understand the commonsense of adding ``y'' to be informal, cute, and childish. 

\begin{wraptable}{r}{8.5cm}
    \vspace{-0.6cm}
    \centering
    \caption{The performance of different methods extracting emotion-cause triplets on ECF.}
    \vspace{0.3cm}
    \scalebox{0.58}{
    \begin{tabular}{l|c|c|c|c|c|c|c|c}
    \toprule
    Method         &   Anger  &  Disgust   &  Fear   & Joy   &  Sadness   &  Surprise   & 6 Avg. & 4 Avg.  \\
    \midrule\midrule
    ECPE-2D\cite{ECPE_2D}    &  25.13 & 0.00 & 0.00 & 41.25 & 21.62 & 43.24 & 30.80 & 33.55  \\
    UECA-Prompt\cite{UECAPrompt} &  27.37 & 12.85 & 7.91 & 37.96 & 22.51 & 39.53 & 30.75 & 32.49  \\
    SHARK\cite{ECTEC_csk}       &  28.65 & 10.42 & 5.33 & 40.41 & 25.35 & 40.45 & 32.24 & 34.33  \\
    \midrule
    EDEN-ECTEC  &  \textbf{40.28} & \textbf{29.79} & \textbf{12.50} & \textbf{49.06} & \textbf{36.40} & \textbf{52.88} & \textbf{43.42} & \textbf{45.13}  \\
    \bottomrule
    \end{tabular}
    }
    \label{ecf_res}
    % \vspace{-0.1cm}
\end{wraptable}

Moving to EDEN-LLaMA, EDEN-LLaMA achieves most of the best results, especially on reasonableness. This demonstrates that EDEN can activate the reasoning ability of LLaMA by LoRA fine-tuning. For cause, EDEN-LLaMA is competitive with SOTA methods on EDEN-DD and is better than SOTA on EDEN-FR, as our Cause Evaluator is not carefully designed like BHG. For emotion, EDEN-LLaMA outperforms SOTA methods on both datasets. To further demonstrate the EDEN explanation can help better emotion recognition, we train an LLaMA2 to directly guess the emotion (LLaMA-7b-EmoGuess). The results emphasize the benefit of EDEN. The above-mentioned outcomes indicate that EDEN can be a substitution for ERD and ECED. For ECED, we provide the result of the Cause Evaluator using golden explanations, which is 75.31 for EDEN-DD and 82.93 for EDEN-FR. This manifests that EDEN still has improving space for ECED using Longformer, and we will design a more sophisticated evaluator in the future. Since EDEN can deal with ERD and ECED, we are curious about whether EDEN extracts better triplets. To this end, we propose a framework called EDEN-ECTEC to perform ECTEC, which is illustrated in Appendix~\ref{frame_work_for_eden_ectec}. Results in Tab.~\ref{ecf_res} show that EDEN is also capable of ECTEC. 

\begin{wraptable}{r}{5.5cm}
\vspace{-0.6cm}
\centering
\caption{The performance of EDEN-LLaMA using different prompts.}
\vspace{0.3cm}
\scalebox{0.65}{
\begin{tabular}{l|l|c|c|c}
\toprule
\multicolumn{2}{l|}{Method}                       & EF       & CF         & Avg.    \\ \midrule\midrule
\multicolumn{5}{c}{EDEN-DD} \\
% \multicolumn{2}{l|}{\multirow{2}{*}{model names}} & \multicolumn{3}{c}{EDEN-DD} \\ \cline{3-5} 
% \multicolumn{2}{c|}{}                             & emotion     & cause         & average   \\ \hline
\multirow{4}{*}{7b}       & full                  & 90.60       & 71.61       & 81.11    \\
                          & speaker-utterance     & 89.03       & 71.98       & 80.51    \\
                          & speaker-turn          & 86.77       & 71.40       & 79.38    \\
                          & speaker               & 90.25       & 70.50       & 80.38    \\
                          & emotion               & 90.50       & 68.76       & 79.63    \\
                          & none                  & 86.41       & 70.22       & 78.32    \\ \midrule
% \multirow{4}{*}{13b}      & full                  & 0.8115       & 0.7009       & 0.7562       \\
%                           & speaker-utterance     & 0.7965       & 0.6957       & 0.7461       \\
%                           & speaker-turn          & 0.8099       & 0.6612       & 0.7356       \\
%                           %& speaker               & 0.7991       & 0.7258       & 0.7625       \\
%                           %& emotion               & 0.8031       & 0.7101       & 0.7566       \\
%                           & none                  & 0.7861       & 0.6784       & 0.7323       \\ \midrule
\multicolumn{5}{c}{EDEN-FR} \\
\multirow{4}{*}{7b}       & full                  & 68.00       & 72.18       & 70.09    \\
                          & speaker-utterance     & 67.42       & 71.27       & 69.35    \\
                          & speaker-turn          & 67.60       & 72.21       & 69.91    \\
                          & speaker               & 67.39       & 70.99       & 69.19    \\
                          & emotion               & 65.82       & 71.69       & 68.76    \\
                          & none                  & 66.59       & 71.18       & 68.89    \\ \bottomrule
\end{tabular}
}
\label{eden_ablation}
\end{wraptable}

Finally, we analyze the performance of variants of EDEN-LLaMA. As shown by EDEN-LLaMA hybrid, more EDEN data can slightly improve most metrics except for Reasonableness. This indicates that LLaMA also learns some co-occurrence features. Besides, the topics, speaker conditions, and dialogue styles in these two datasets are of evident difference, which may introduce some gaps to LLaMA's reasoning learning. As for EDEN-LLaMA sup, the improvement in EDEN-FR is more evident. This indicates that external commonsense knowledge can help predict emotion and cause, which aligns with previous works~\cite{COSMIC,SKAIG,KEC,KBCIN,Bipartite}. However, commonsense pretraining shows no improvement in reasonableness. The reason may be that these commonsense data only support single-step reasoning. Reasoning speakers' inner activities is more complicated as multi-hop or compositional reasoning may be required.

\begin{table}
\centering
\caption{The performance of different methods when directly generating the explanation.}
\scalebox{0.68}{
\begin{tabular}{l|c|c|c|c|c|c|c||c|c|c|c|c|c|c}
\toprule
\multirow{2}{*}{Method}     & \multicolumn{7}{c||}{EDEN-DD} & \multicolumn{7}{c}{EDEN-FR}\\\cline{2-15}
                      & B-3    & B-4    & R-L    & MT     & Cr     & EF & CF & B-3    & B-4    & R-L    & MT     & Cr     & EF & CF \\ 
\hline
BART-lg            & 14.93 & 10.01 & 27.25 & 32.68 & 12.42 & 82.85 & 65.51 & 16.93 & 10.29 & 23.07 & 26.25 & 17.04 & 55.94 & 70.62 \\
GPT2-lg            & 12.87 & 8.63 & 26.87 & 29.55 & 11.87 & 82.94 & 64.36 & 13.50 & 8.23 & 21.67 & 22.65 & 14.79 & 51.25 & 65.32 \\ \hline
% ChatGPT emotion first & 0.0660 & 0.0389 & 0.2094 & 0.3365 & 0.0048 &    &    &    \\
% ChatGPT 0-shot        & 0.0904 & 0.0512 & 0.2596 & 0.3782 & 0.0330 & 0.7636 & 0.6299 &    \\
% ChatGPT 1-shot        & 9.76 & 5.82 & 26.05 & 38.07 & 2.35 & 77.51 & 64.59 \\
ChatGPT-5s       & 11.53 & 7.26 & 28.22 & 40.04 & 5.28 & 79.57 & 65.87 & 13.53 & 7.89 & 23.97 & 34.35 & 5.02 & 61.69 & 74.71 \\ \hline
E-LLaMA-7b         & 17.18 & 11.69 & 30.02 & 33.72 & 16.96 & 81.81 & 66.56 & 18.52 & 11.40 & 24.21 & 26.57 & 16.71 & 59.10 & 73.31 \\
E-LLaMA-13b        & 17.88 & 12.16 & 30.20 & 34.87 & 16.11 & 86.39 & 66.40 & 19.43 & 12.07 & 24.97 & 27.88 & 19.77 & 62.20 & 74.15 \\
% EDEN-LLaMA-7b hybrid  & 0.1771 & 0.1198 & 0.3027 & 0.3452 & 0.1695 & 0.8487 & 0.6729 \\
% EDEN-LLaMA-13b hybrid & 0.1772 & 0.1202 & 0.3031 & 0.3437 & 0.1725 & 0.8623 & 0.6816 \\ 
\bottomrule
\end{tabular}
}
\label{eden_dd_fr_2}
\vspace{-0.5cm}
\end{table}

\textbf{Directly generating the explanation.}~~~~The explanation $\mathcal{E}$ in EDEN acts as a CoT for emotion recognition in the form of ``The emotion is $e_t$. $\mathcal{E}$''. In this part, we let models only generate $\mathcal{E}$, which is equivalent to the CoT going first and then predicting emotion. The results are shown in Tab.~\ref{eden_dd_fr_2}. It can be seen that both the prediction and the explanation are affected, especially to PLMs and EDEN-LLaMA. The reason may be the erroneous accumulation (which is also said to be hallucination by~\cite{Multomodal-CoT}). The findings align with the study of CoT in ScienceQA~\cite{ScienceQA}. Compared to the emotion, the impact on the cause is less because the cause is generated first. The drop in CF also shows the mutual indication of emotion and cause. It is worth noting that ChatGPT is less affected. This demonstrates its consistency in deducing emotion as it is one of the most powerful LLMs. 

% and even behaves slightly better in EDEN-FR

\textbf{Instruction Ablation.}~~~~The instruction for EDEN-LLaMA (denoted as ``\textit{full}'') considers several factors: the emotion list, the target speaker, the turn id of the target utterance, and the target utterance. We ablate these factors by removing the turn id (``\textit{speaker-utterance}''); removing the target utterance (``\textit{speaker-turn}''); removing the target utterance and its turn id (``\textit{speaker}''); only keeping the emotion list (``\textit{emotion}''); removing all factors (``\textit{none}''). The performance of EDEN-LLaMA-7b on dev sets is illustrated in Tab.~\ref{eden_ablation}. The results show that considering the speaker, the target utterance, and the emotion list is important as these factors can guide models to realize the carrier of the emotion. Besides, if EDEN-LLaMA is not told the emotion range, it acts like ChatGPT to predict emotions out of the emotion list. 

% \subsection{Experimental Settings}

\textbf{Error Analysis.}~~~~We study the error case based on the most intuitive factor: emotion. By collecting samples that both ChatGPT 5-shots and EDEN-LLaMA-13b make the same error prediction, we can obtain confusion matrices of emotion on EDEN-DD and EDEN-FR, which is illustrated in Fig.~\ref{heatmap}, Appendix~\ref{appendix_error}. It is shown that negative emotions are confused with each other and ``surprise'' is confused with all other emotions. The reason for the case of ``surprise'' is that ``surprise'' naturally has any valence, which means it can be positive or negative~\cite{EVT}. As the distinction between emotions with the same valence is subtle, it requires more refined reasoning to appraisal~\cite{appraisal}. Besides, emotions with different valences (e.g., ``happiness'' and ``anger'') also show confusion. We utilize the pysentiment tool to compute the polarity score of the target utterance $\epsilon_{u}$ and its dialogue $\epsilon_{d}$. For positive utterances, $\epsilon_{u}$=$0.08_{(\pm 0.55)}$ and $\epsilon_{c}$=$0.01_{(\pm 0.45)}$. For negative utterances, $\epsilon_{u}$=$0.35_{(\pm 0.60)}$ and $\epsilon_{c}$=$0.28_{(\pm 0.31)}$. This indicates that these samples contain expressions whose literal sentiment is opposite to the target emotion, and models tend to utilize this shallow semantic information. By viewing these samples, we found that part of the reasons may come from models' to-be-strengthened ability to compositional and multi-hop reasoning. As shown in Fig.~\ref{eden_cases} (c), the dialogue contains negative expressions like ``bad neighborhood''. Models depend on them regardless of the compositional realization of the cozy atmosphere in the context and the true intents of Phoebe and Gary. 

% ``gunshot'' and

\section{Conclusion and Limitations}

In this work, we propose a new task called ``Emotion Deducing Explanation in Dialogues'' (EDEN \mycustomsymbol) to thoroughly and interpretably evaluate models' ability to find emotional triggers and rationally reason the process of these triggers stimulating the emotion. To support EDEN, we curated two datasets EDEN-DD and EDEN-FR by human efforts. We further adapted experiments with a range of models on these datasets and found that LLMs are more competent on EDEN and their reasoning ability can be activated by EDEN to achieve better emotion and cause recognition. EDEN provides a new research direction of explainable emotion understanding in dialogues, which can replace previous counterparts in the era of LLMs. 

\textbf{Limitations.}~~~~The huge human labor of EDEN \mycustomsymbol\ makes its size hard to scale. Besides, we only annotate emotional utterances due to the burden and there are considerable \textit{neutral} utterances without explanation. In future works, we will explore more effective annotating methods. 

\clearpage

\appendix

\section{Prompt Templates Used in Our Experiments}\label{appendix_prompt}

\begin{figure*}[]
\begin{lstlisting}[language=prompt]

***[SYSTEM]***
You are given a dialogue between speaker A and speaker B. The last utterance is a targeted utterance showing a kind of emotion. Some of the former contextual utterances of the targeted utterance or the targeted utterance itself are labeled as the causal utterances to the emotion of the targeted utterance. Your goal is to explain why these labeled utterances are the causes.

***[USER]***
Given the dialogue:

(1) |||{s1}|||: |||{u1}|||
...
(|||n|||) |||{sn}|||: |||{un}||| 

In the dialogue, (|||n|||) is the targeted utterance showing the emotion of |||{target_emotion}|||. (|||i|||), ..., (|||j|||) are labeled as the causal utterances. Please explain step-by-step why (|||i|||), ..., (|||j|||) are the cause to speaker |||{sn}|||'s |||{target_emotion}||| at (|||n|||) based on the dialogue.

***[Assistant]***
///The detailed explanation generated by ChatGPT.///

***[USER]***
Please summarize your explanation.
\end{lstlisting}
\caption{The two-round prompt template for ChatGPT to generate the original explanation with the given gold emotion and causes. The example is for DailyDialogue. }
\label{two_round}
\end{figure*}

\begin{figure*}[]
\begin{lstlisting}[language=prompt]
***[SYSTEM]***
Please analyze how the emotional causes lead to a speaker's emotion in a dialogue. 

***[USER]***
Give a dialogue clip where a speaker is expressing a kind of emotion at the last utterance. You should identify the emotion of the speaker and explain how the emotional triggers lead to the emotion. For the explanation, you should firstly find the emotion-triggering causes, which can be events happened in the dialogue, sensory input, or phraseology of the speaker. Then, you should explain why these emotional causes can lead to the speaker's possible emotion by analyzing speakers' mind-thinking and psychological activities or reasoning with commonsense. Lastly, determine the emotion based on your explanation about the causes' effects. There are 6 types of emotions can be guessed at: |||[happiness, sadness, fear, surprise,||| |||anger, disgust]|||. Your explanation should be formed as a paragraph of text, and the number of words in your explanation should be around 50-200.

Below are some examples that explain how the causes deduce the emotion:

///[example 1]///
...
///[example 5]///
====
Now, give the dialogue clip below:
///{dialogue_clip}///

In the dialogue clip, the speaker |||{sn}||| is showing a kind of emotion at the last utterance (|||n|||), whose content is |||{tn}|||. You should identify the emotion and explain how the emotion is evoked by the dialogue clip. From the emotion-triggering events or factors in the dialogue to the target emotion of the speaker |||{sn}|||, your answer is:
\end{lstlisting}
\caption{Prompt template for ChatGPT 5-shots.}
\label{chatgpt_prompt}
\end{figure*}

\begin{figure}[]
\begin{lstlisting}[language=prompt]
***[SYSTEM]***
Explain the process of a speaker's emotion arising in a dialogue.

***[USER]***
Provide a dialogue clip below:

///{dialogue_clip}///

In the dialogue clip, |||{target_speaker}||| expresses a non-neutral emotion at the last utterance (|||{last_turn}|||), where the last utterance (|||{last_turn}|||) is "|||{last_utterance}|||". Your goal is to identify the emotion of |||{target_speaker}||| at (|||{last_turn}|||) and explain how the dialogue evokes the emotion. Your explanation should follow such a format: Firstly, identify the potential events or factors triggering this emotion. Subsequently, infer how these events or factors elicit the emotion through common-sense reasoning or an analysis of the speaker's psychological activities. Finally, determine the type of emotion induced by these evident triggers. The emotion of |||{||| |||target_speaker}||| at (|||{last_turn}|||) can be |||"sadness", "fear", "disgust", "happiness", "anger", or "surprise"|||. 
Your answer is: 
\end{lstlisting}
\caption{Prompt template for EDEN-LLaMA, which is called ``full''.}
\label{eden_llama_prompt}
\end{figure}

In our work, ChatGPT and LLaMA are the main models for assistance and experiments. Due to the limited space in the main body, we illustrate the prompt templates we used in this appendix chapter. 

To assist the annotation process, ChatGPT is used to generate the original analysis based on the gold emotion and causes. The two-round prompt template is shown in Fig.~\ref{two_round}, where \{sn\} is a speaker; \{un\} is \{sn\}'s utterance at the n-th turn; \{target\_emotion\} is the gold emotion; (i),...,(j) are the ids of gold causal utterances. 

ChatGPT uses in-context learning to generate EDEN \mycustomsymbol\ explanation. Fig.~\ref{chatgpt_prompt} illustrates the prompt template for ChatGPT 5-shots. In the figure, [example 1] is a demonstration containing its dialogue clip and gold explanation; \{sn\} is the target speaker; \{tn\} is the utterance at the n-th turn. 

EDEN-LLaMA performs instruct tuning using the prompt template in Fig.~\ref{eden_llama_prompt}, which is denoted as ``\textit{full}''. As mentioned in the main body, the ``\textit{full}'' prompt considers several factors: the target speaker \{target\_speaker\}; their utterance \{last\_utterance\} at the \{last\_turn\}-th turn; the emotion list [``sadness'', ``fear'', ``disgust'', ``happiness'', ``anger'', ``surprise'']. If \{last\_turn\} is removed, the target speaker and their utterance are kept, which is denoted as the ``\textit{speaker-utterance}'' prompt. If \{last\_utterance\} is removed, the target speaker and the turn id of the last utterance are kept, which is denoted as the ``\textit{speaker-turn}'' prompt. If both \{last\_utterance\} and \{last\_turn\} are removed, the target speaker is still kept and models need to learn the last utterance by words ``the last utterance'' in the prompt, which is denoted as the ``\textit{speaker}'' prompt. If \{last\_utterance\}, \{last\_turn\}, and \{last\_speaker\} are removed, only the emotion list remains, which is denoted as the ``\textit{emotion}'' prompt. Removing all these factors, the prompt is called ``\textit{none}''. 

The prompt template for Reasonableness is shown in Fig.~\ref{reasonableness_prompt}. In the prompt, detailed criteria for the scoring are colored rose red. It can be seen that only correctly predicting the emotion and causes is not enough to get a high score, which sets a requirement of properly and rationally commonsense reasoning. 

\begin{figure}[]
\begin{lstlisting}[language=prompt]
***[SYSTEM]***
You are a quality rater

***[USER]***
Your task is to rate the explanation on one metric. Please make sure you read and understand these instructions carefully. Please keep this dialogue open while reviewing, and refer to it as needed. 

Evaluation Criteria: Reasonableness (1-3) Does the input explanation generated by a machine serve as a reasonable and correct analysis to how the emotion of a target speaker is deduced from the dialogue? 
<<<- A score of 1 (no) can be that the input explanation makes a wrong emotion prediction. 
- A score of 1 (no) can be that the input explanation makes a right emotion prediction but finds a wrong emotional causes from the dialogue. 
- A score of 1 (no) can be that the input explanation makes a right emotion prediction but wrongly analyzes how the emotional causes lead to the emotion by making factual errors or improperly commonsense reasoning. Besides, the explanation is lengthy and redundant. 
- A score of 2 (somewhat) means that the input explanation can guess the right emotion, finds partial emotional causes from the dialogue, and partially makes an acceptable analysis of why the emotional causes lead to the emotion in a commonsense way, but still has some improper deductions and need improvement according to the potential explanation. 
- A score of 3 (yes) means that the input explanation can start from the right and accurate emotion causes and properly deduce how the emotional causes lead to the right emotion by analyzing speakers' psychological activities well and showing human commonsense. You can refer to the potential explanation for demonstration. >>>

Evaluation Steps: 
1. Read the conversation. Read the target speaker and the target emotion. 
2. Read the potential explanation for analyzing how the emotional causes in the dialogue lead to the target speaker's emotion. 
3. Evaluate the Reasonableness of the input explanation based on the given dialogue. 
4. Assign a rating score of 1, 2, or 3 for Reasonableness based on the evaluation criteria. 

Question: By reading the dialogue and the potential explanation which is a good demonstration, does the input explanation serve as a reasonable and correct analysis to how the emotion |||{en}||| of |||{sn}||| at (|||n|||) is deduced from the dialogue? (On a scale of 1-3, with 1 meaning the explanation is logically erroneous and 3 meaning the explanation is reasonable) 

Dialogue: 
(1) |||{s1}|||: |||{u1}|||
...
(|||n|||) |||{sn}|||: |||{un}||| 

Potential explanation: 
|||{gold_explanation}|||

Input explanation: 
|||{generated_explanation}|||

Evaluation Form (Answer by starting with "Analysis:" to analyze the given example regarding the evaluation criteria as concise as possible, and then give the numeric rating on the next line by "Rating:"): 

- Reasonableness:
\end{lstlisting}
\caption{Prompt template for Reasonableness. The text with {\color{frenchrose}rose color} is the detailed criteria of Reasonableness.}
\label{reasonableness_prompt}
\end{figure}

\section{Implementation Details and Hyper-Parameters}\label{implement}

For the training details, generative models are fine-tuned with PyTorch distributed data-parallel mode with mixed precision of BFloat16 and TensorFloat32. The batch size per device is set to 4 and the world size is set to 4. The number of epochs is set to 5. For fully fine-tuned BART and GPT2, the learning rate is set to 1e-5 as usual and the learning rate of LoRA is set to 1e-4. For LoRA fine-tuning, the rank $r$ is set to 8; the alpha is set to 16; the dropout rate is set to 0.05; LoRA adaptors are attached to the query, key, and value parameters; the backbone is frozen to int8. As the world size is 4 and BF16/TF32 is used, we use 4 pieces of A100 GPU whose memory size is 40G for experiments. Of course, fewer computation resources also support the training. 

For complimentary commonsense pretraining, ATOMC~\cite{ATOMIC}, GLUCOSE~\cite{GLUCOSE}, and CICEROv2~\cite{CICEROv2} are utilized as the training data. In ATOMIC, the knowledge of \textit{oEffect}, \textit{oReact}, \textit{oWant}, \textit{xAttr}, \textit{xEffect}, \textit{xIntent}, \textit{xReact}, \textit{xReason}, \textit{xWant}, \textit{Causes}, \textit{Desires}, \textit{HasProperty}, and  \textit{NotDesires} are used, which contributes 550424 items of data. In GLUCOSE, all the 10 types of knowledge about \textit{Causes/Enables}, \textit{Motivates}, \textit{Results-in}, which contributes 304099 items of data. In CICEROv2, all the mutual data are used, which contributes 5495 items of data. The complimentary commonsense pretraining sets the batch size per device to 8 using accumulation steps of 4 and trains for 4 epochs. Other hyper-parameters for it are the same as those in the last paragraphs. As we mentioned in the main body, we set a prompt template pool to transfer the knowledge into natural language. Due to the limited space for the demonstration, we show the transferring in our code. 

The temperature is set to 1.0 for ChatGPT and GPT4 for generations. For other generative models, the temperature is set to 0.2 adding a repetition penalty of 1.2. The maximum number of new tokens is 200. TFD~\cite{TFD} and BHG~\cite{Bipartite} are re-implemented based on their official code\footnote{\url{https://github.com/TuGengs/TFD}}\footnote{\url{https://github.com/SteveKGYang/BHG}} using a piece of 40G A100 GPU. 

\section{EDEN for Emotion-Cause Triplet Extraction}\label{frame_work_for_eden_ectec}

To achieve emotion-cause triplets extraction in dialogues using EDEN~\mycustomsymbol, there are two points that need to be solved: (1) EDEN~\mycustomsymbol only deals with samples whose last utterance is in an unknown emotion and skips those dialogues whose last utterance is neutral. The reason for this is that neutral utterances appear frequently in dialogues and our human labor cannot cover such huge burdens. (2) EDEN~\mycustomsymbol generates natural text instead of triplets in numbers. 

\begin{figure}
    \centering
    \includegraphics[width=0.95\textwidth]{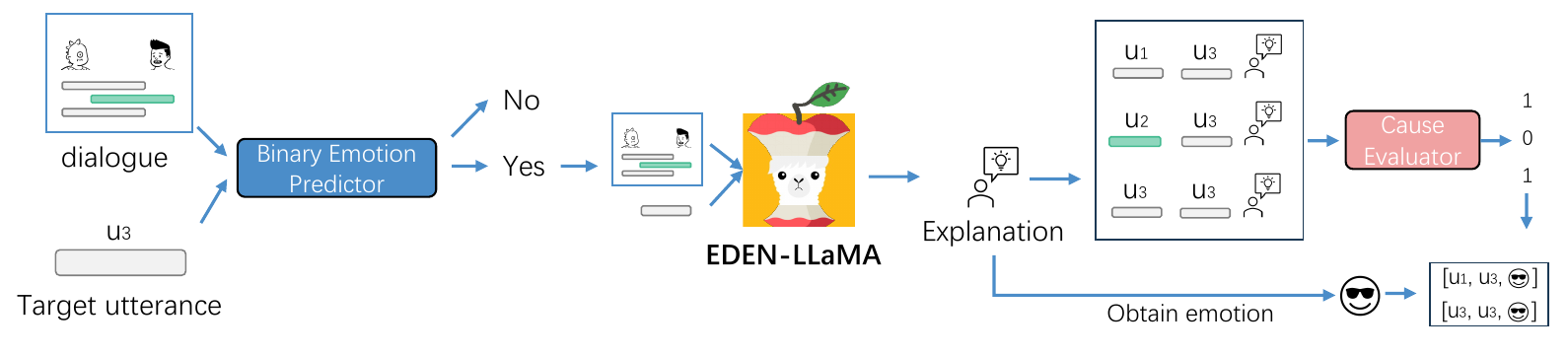}
    \caption{The framework of EDEN-ECTEC to extract emotion-cause triplets from dialogues. }
    \label{eden_ectec}
\end{figure}

For point 1, we propose to train a binary emotion predictor $\mathcal{B}$. For every dialogue clip, inputting the clip and its last utterance, $\mathcal{B}$ identifies whether the last utterance possesses an emotion. If it has an emotion, the label will be 1, otherwise, the label will be 0 to indicate neutral. Samples predicted to be 1 will be input to EDEN-LLaMA-7b for EDEN~\mycustomsymbol analysis. To solve point 2, we utilize our Cause Evaluator to guess causal utterances for the last utterance. Since the emotion is easy to extract from the explanation, the final triplets can be constructed. We denote the framework as EDEN-ECTEC, whose flow chart is depicted in Fig.~\ref{eden_ectec}. 

\section{Additional Error Cases}\label{appendix_error}

\begin{figure}[h]
    \centering
    \includegraphics[width=0.8\textwidth]{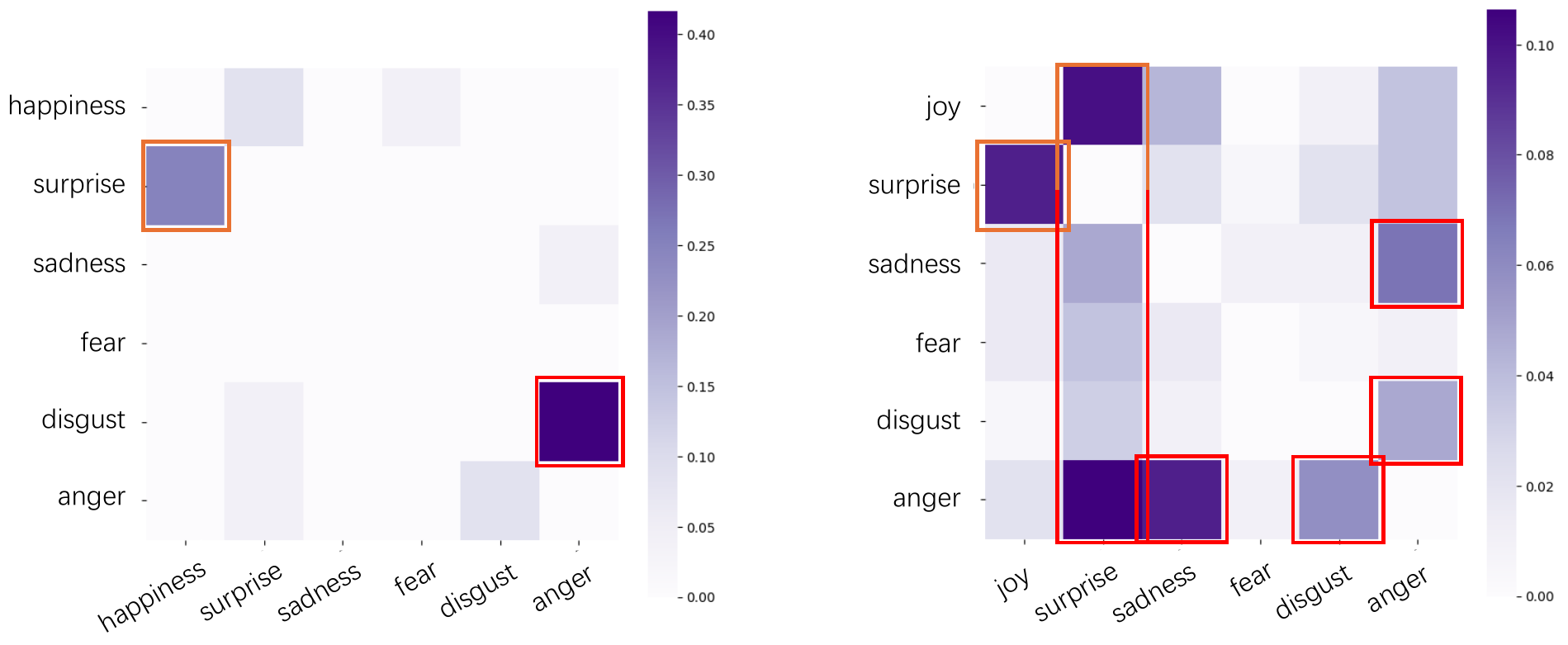}
    \caption{Confusion matrices of emotions calculated from the same erroneous samples of ChatGPT 5-shots and EDEN-LLaMA-13b. The left one is for EDEN-DD and the right one is for EDEN-FR. }
    \label{heatmap}
\end{figure}

We study the error cases by collecting samples on that both ChatGPT 5-shots and EDEN-LLaMA-13b make the same and wrong emotion prediction. Fig.~\ref{heatmap} provides two confusion matrices, where the left one is for EDEN-DD and the right one is for EDEN-FR. Both matrices have removed their diagonal. The blocks with darker colors are emphasized to show that confusion frequently happens (1) between negative emotions; and (2) between ``surprise'' and all other emotions. By the way, the pysentiment tool is in this url\footnote{\url{https://github.com/hanzhichao2000/pysentiment}}. 

\section{The Annotating Interface}\label{appendix_interface}

\begin{figure}[H]
    \centering
    \includegraphics[width=0.98\textwidth]{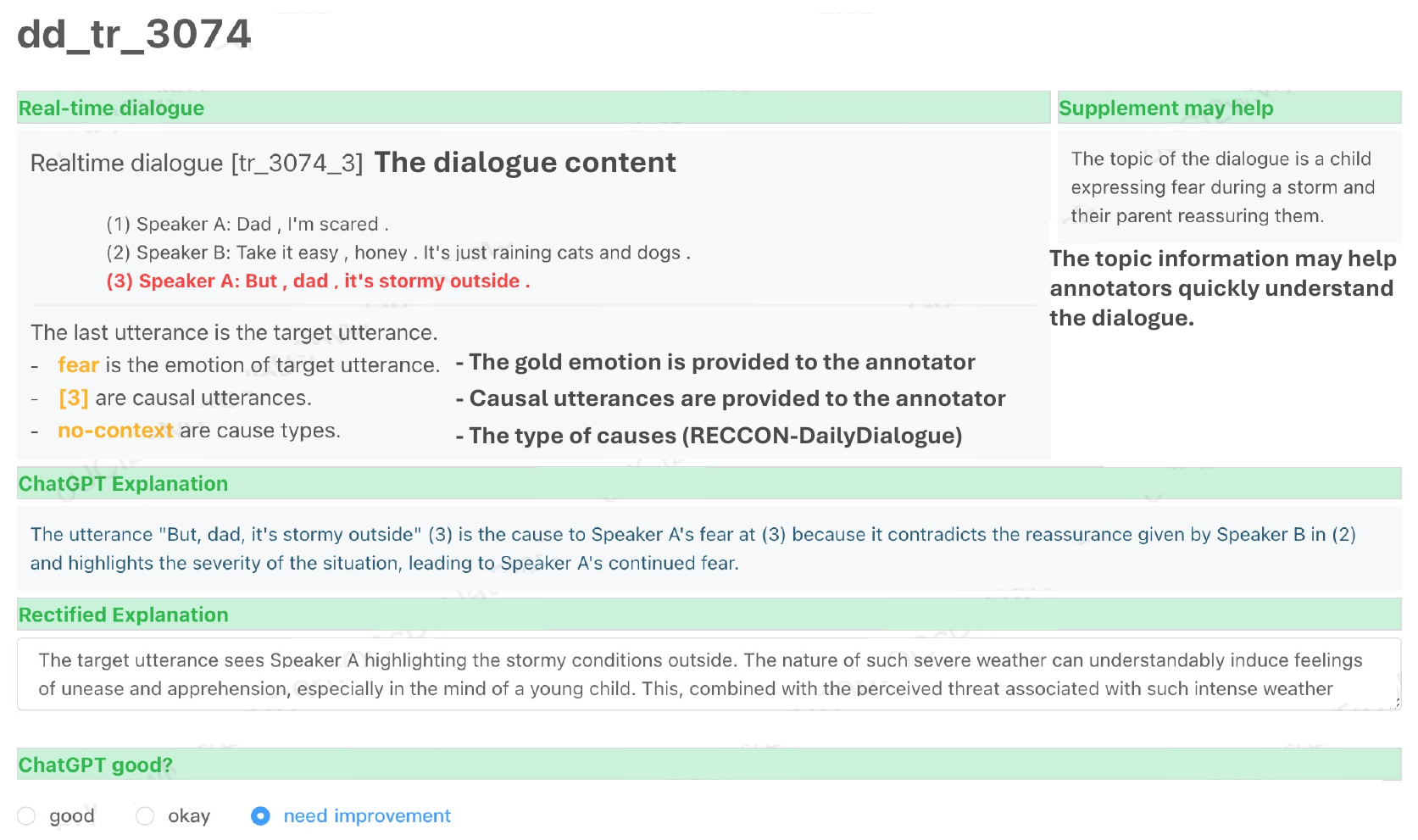}
    \caption{The annotating interface screenshot.}
    \label{screenshot}
\end{figure}

The annotating interface screenshot is shown in Fig.~\ref{screenshot}, where all the gold labels and additional topic information are given. The annotators are required to read the ChatGPT explanation, score it, and make revisions accordingly. 

\section{More Illustrations of Model Outcomes}

\begin{figure}[H]
    \centering
    \includegraphics[width=0.8\textwidth]{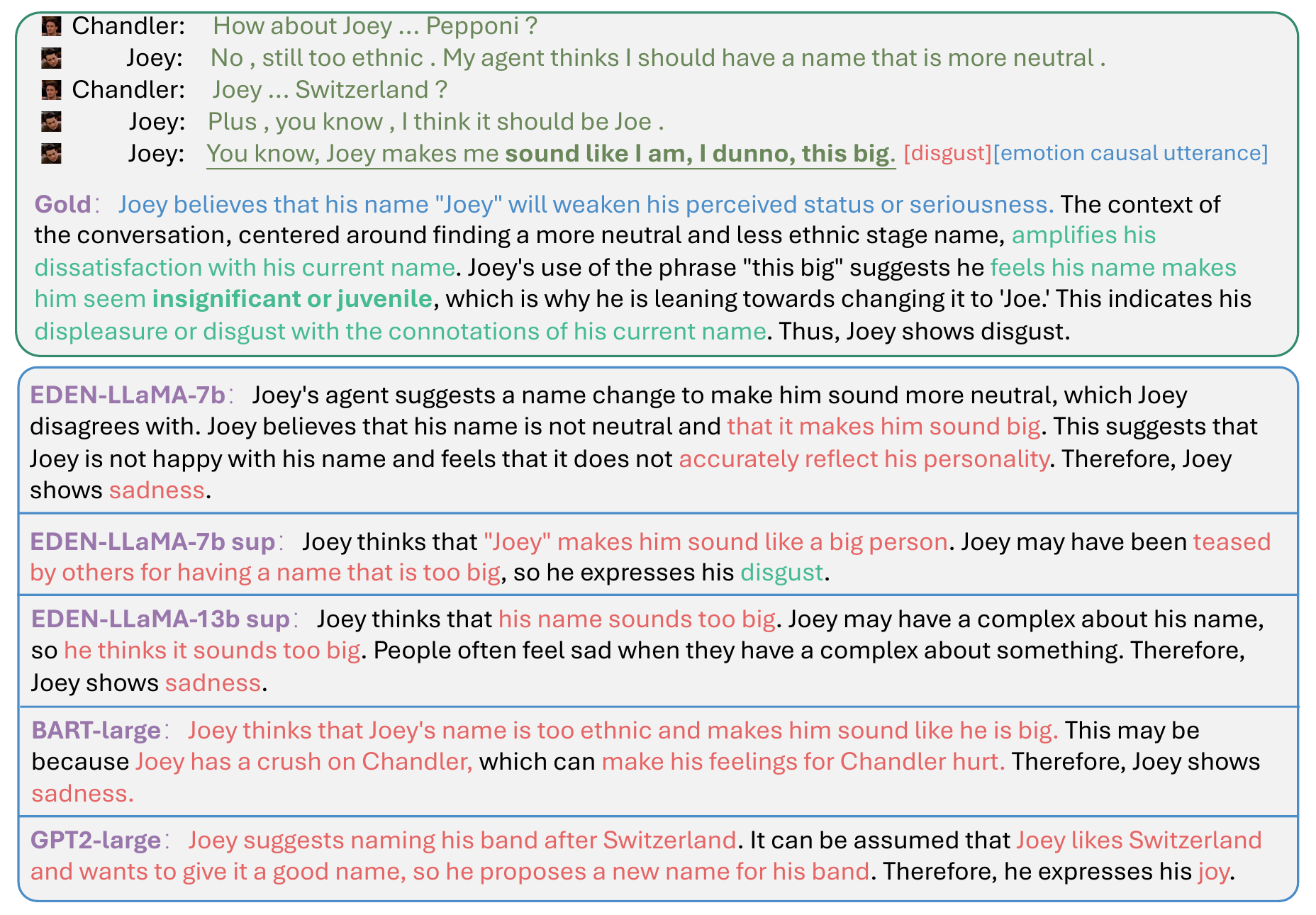}
    \caption{Other models on the case of ``Joey changing name'' in Fig.~\ref{eden_cases}~(b). }
    \label{appendix_case_1}
\end{figure}

\begin{figure}[H]
    \centering
    \includegraphics[width=0.8\textwidth]{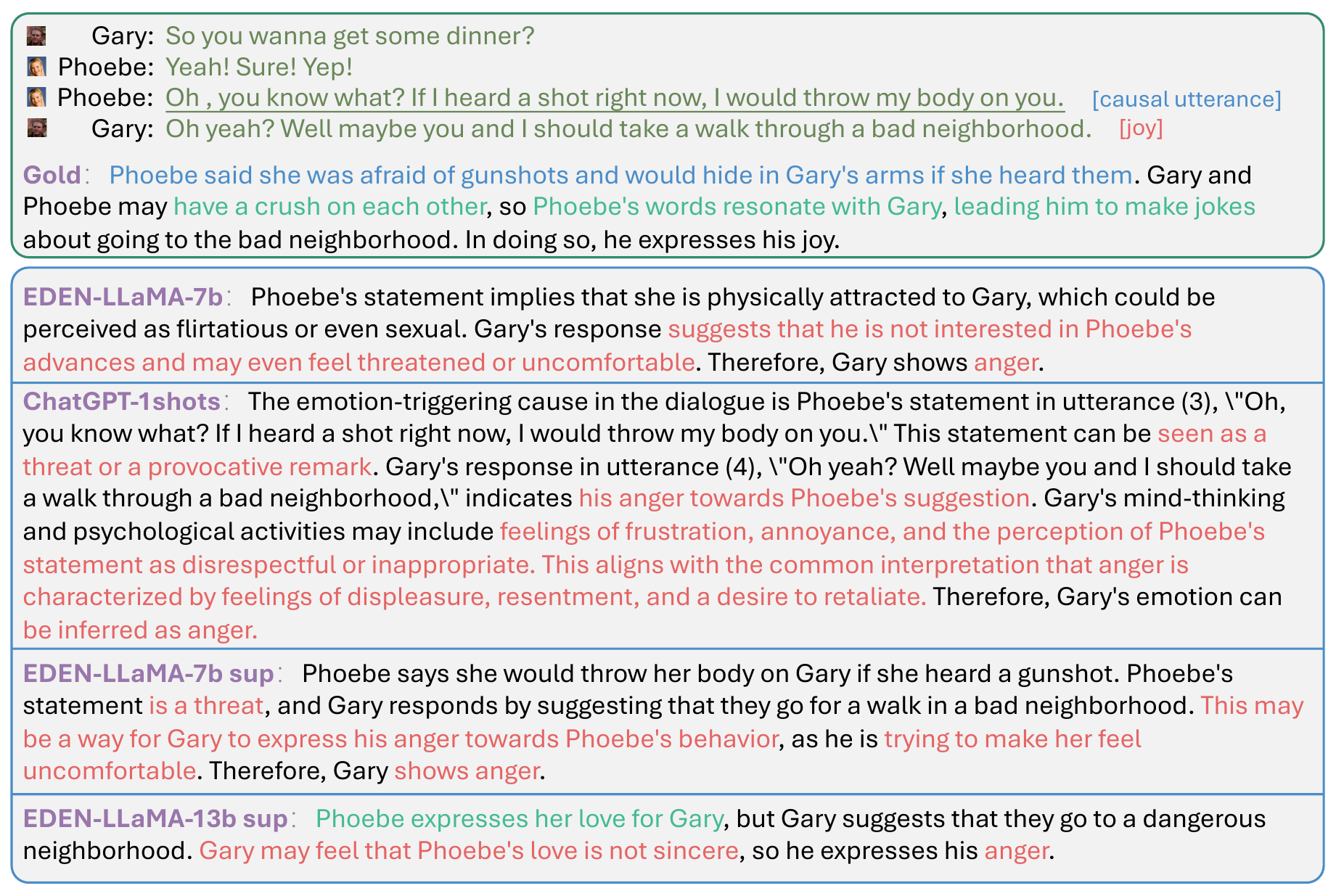}
    \caption{Other models on the case of ``Phoebe and Gary flirting'' in Fig.~\ref{eden_cases}~(c). }
    \label{appendix_case_2}
\end{figure}

More illustrations of the outcomes of models to the cases in Fig~\ref{eden_cases} (b)(c) are shown in Fig.~\ref{appendix_case_1} and Fig.~\ref{appendix_case_2}. In these figures, models fail to make a promising analysis due to the to-be-strengthened reasoning ability. 

\begin{thebibliography}{10}

\bibitem{aronson2005social}
Elliot Aronson, Timothy~D Wilson, and Samuel~R Sommers.
\newblock {\em Social psychology}.
\newblock Pearson Education India, 2005.

\bibitem{COMET}
Antoine Bosselut, Hannah Rashkin, Maarten Sap, Chaitanya Malaviya, Asli Celikyilmaz, and Yejin Choi.
\newblock {COMET:} commonsense transformers for automatic knowledge graph construction.
\newblock In Anna Korhonen, David~R. Traum, and Llu{\'{\i}}s M{\`{a}}rquez, editors, {\em Proceedings of the 57th Conference of the Association for Computational Linguistics, {ACL} 2019, Florence, Italy, July 28- August 2, 2019, Volume 1: Long Papers}, pages 4762--4779. Association for Computational Linguistics, 2019.

\bibitem{GPT3}
Tom~B. Brown, Benjamin Mann, Nick Ryder, Melanie Subbiah, Jared Kaplan, Prafulla Dhariwal, Arvind Neelakantan, Pranav Shyam, Girish Sastry, Amanda Askell, Sandhini Agarwal, Ariel Herbert{-}Voss, Gretchen Krueger, Tom Henighan, Rewon Child, Aditya Ramesh, Daniel~M. Ziegler, Jeffrey Wu, Clemens Winter, Christopher Hesse, Mark Chen, Eric Sigler, Mateusz Litwin, Scott Gray, Benjamin Chess, Jack Clark, Christopher Berner, Sam McCandlish, Alec Radford, Ilya Sutskever, and Dario Amodei.
\newblock Language models are few-shot learners.
\newblock In Hugo Larochelle, Marc'Aurelio Ranzato, Raia Hadsell, Maria{-}Florina Balcan, and Hsuan{-}Tien Lin, editors, {\em Advances in Neural Information Processing Systems 33: Annual Conference on Neural Information Processing Systems 2020, NeurIPS 2020, December 6-12, 2020, virtual}, 2020.

\bibitem{EVT}
Judee~K Burgoon and Stephen~B Jones.
\newblock Toward a theory of personal space expectations and their violations.
\newblock {\em Human communication research}, 2(2):131--146, 1976.

\bibitem{IEMOCAP}
Carlos Busso, Murtaza Bulut, Chi{-}Chun Lee, Abe Kazemzadeh, Emily Mower, Samuel Kim, Jeannette~N. Chang, Sungbok Lee, and Shrikanth~S. Narayanan.
\newblock {IEMOCAP:} interactive emotional dyadic motion capture database.
\newblock {\em Lang. Resour. Evaluation}, 42(4):335--359, 2008.

\bibitem{GoEmotions}
Dorottya Demszky, Dana Movshovitz{-}Attias, Jeongwoo Ko, Alan~S. Cowen, Gaurav Nemade, and Sujith Ravi.
\newblock Goemotions: {A} dataset of fine-grained emotions.
\newblock In Dan Jurafsky, Joyce Chai, Natalie Schluter, and Joel~R. Tetreault, editors, {\em Proceedings of the 58th Annual Meeting of the Association for Computational Linguistics, {ACL} 2020, Online, July 5-10, 2020}, pages 4040--4054. Association for Computational Linguistics, 2020.

\bibitem{SarcasmExplanation}
Poorav Desai, Tanmoy Chakraborty, and Md.~Shad Akhtar.
\newblock Nice perfume. how long did you marinate in it? multimodal sarcasm explanation.
\newblock In {\em Thirty-Sixth {AAAI} Conference on Artificial Intelligence, {AAAI} 2022, Thirty-Fourth Conference on Innovative Applications of Artificial Intelligence, {IAAI} 2022, The Twelveth Symposium on Educational Advances in Artificial Intelligence, {EAAI} 2022 Virtual Event, February 22 - March 1, 2022}, pages 10563--10571. {AAAI} Press, 2022.

\bibitem{ECPE_2D}
Zixiang Ding, Rui Xia, and Jianfei Yu.
\newblock {ECPE-2D:} emotion-cause pair extraction based on joint two-dimensional representation, interaction and prediction.
\newblock In Dan Jurafsky, Joyce Chai, Natalie Schluter, and Joel~R. Tetreault, editors, {\em Proceedings of the 58th Annual Meeting of the Association for Computational Linguistics, {ACL} 2020, Online, July 5-10, 2020}, pages 3161--3170. Association for Computational Linguistics, 2020.

\bibitem{ekman}
Paul Ekman.
\newblock Facial expression and emotion.
\newblock {\em American psychologist}, 48(4):384, 1993.

\bibitem{ERC_Survey}
Yao Fu, Shaoyang Yuan, Chi Zhang, and Juan Cao.
\newblock Emotion recognition in conversations: A survey focusing on context, speaker dependencies, and fusion methods.
\newblock {\em Electronics}, 12(22):4714, 2023.

\bibitem{COSMIC}
Deepanway Ghosal, Navonil Majumder, Alexander~F. Gelbukh, Rada Mihalcea, and Soujanya Poria.
\newblock {COSMIC:} commonsense knowledge for emotion identification in conversations.
\newblock In Trevor Cohn, Yulan He, and Yang Liu, editors, {\em Findings of the Association for Computational Linguistics: {EMNLP} 2020, Online Event, 16-20 November 2020}, volume {EMNLP} 2020 of {\em Findings of {ACL}}, pages 2470--2481. Association for Computational Linguistics, 2020.

\bibitem{CICERO}
Deepanway Ghosal, Siqi Shen, Navonil Majumder, Rada Mihalcea, and Soujanya Poria.
\newblock {CICERO:} {A} dataset for contextualized commonsense inference in dialogues.
\newblock In Smaranda Muresan, Preslav Nakov, and Aline Villavicencio, editors, {\em Proceedings of the 60th Annual Meeting of the Association for Computational Linguistics (Volume 1: Long Papers), {ACL} 2022, Dublin, Ireland, May 22-27, 2022}, pages 5010--5028. Association for Computational Linguistics, 2022.

\bibitem{LoRA}
Edward~J. Hu, Yelong Shen, Phillip Wallis, Zeyuan Allen{-}Zhu, Yuanzhi Li, Shean Wang, Lu~Wang, and Weizhu Chen.
\newblock Lora: Low-rank adaptation of large language models.
\newblock In {\em The Tenth International Conference on Learning Representations, {ICLR} 2022, Virtual Event, April 25-29, 2022}. OpenReview.net, 2022.

\bibitem{ECR_Chain}
Zhaopei Huang, Jinming Zhao, and Qin Jin.
\newblock Ecr-chain: Advancing generative language models to better emotion-cause reasoners through reasoning chains, 2024.

\bibitem{COMET2020}
Jena~D. Hwang, Chandra Bhagavatula, Ronan~Le Bras, Jeff Da, Keisuke Sakaguchi, Antoine Bosselut, and Yejin Choi.
\newblock (comet-) atomic 2020: On symbolic and neural commonsense knowledge graphs.
\newblock In {\em Thirty-Fifth {AAAI} Conference on Artificial Intelligence, {AAAI} 2021, Thirty-Third Conference on Innovative Applications of Artificial Intelligence, {IAAI} 2021, The Eleventh Symposium on Educational Advances in Artificial Intelligence, {EAAI} 2021, Virtual Event, February 2-9, 2021}, pages 6384--6392. {AAAI} Press, 2021.

\bibitem{bird_cannot_fly}
Nora Kassner and Hinrich Sch{\"{u}}tze.
\newblock Negated and misprimed probes for pretrained language models: Birds can talk, but cannot fly.
\newblock In Dan Jurafsky, Joyce Chai, Natalie Schluter, and Joel~R. Tetreault, editors, {\em Proceedings of the 58th Annual Meeting of the Association for Computational Linguistics, {ACL} 2020, Online, July 5-10, 2020}, pages 7811--7818. Association for Computational Linguistics, 2020.

\bibitem{SarcasmExplanationInDialogue}
Shivani Kumar, Atharva Kulkarni, Md.~Shad Akhtar, and Tanmoy Chakraborty.
\newblock When did you become so smart, oh wise one?! sarcasm explanation in multi-modal multi-party dialogues.
\newblock In Smaranda Muresan, Preslav Nakov, and Aline Villavicencio, editors, {\em Proceedings of the 60th Annual Meeting of the Association for Computational Linguistics (Volume 1: Long Papers), {ACL} 2022, Dublin, Ireland, May 22-27, 2022}, pages 5956--5968. Association for Computational Linguistics, 2022.

\bibitem{BART}
Mike Lewis, Yinhan Liu, Naman Goyal, Marjan Ghazvininejad, Abdelrahman Mohamed, Omer Levy, Veselin Stoyanov, and Luke Zettlemoyer.
\newblock {BART:} denoising sequence-to-sequence pre-training for natural language generation, translation, and comprehension.
\newblock In Dan Jurafsky, Joyce Chai, Natalie Schluter, and Joel~R. Tetreault, editors, {\em Proceedings of the 58th Annual Meeting of the Association for Computational Linguistics, {ACL} 2020, Online, July 5-10, 2020}, pages 7871--7880. Association for Computational Linguistics, 2020.

\bibitem{SKAIG}
Jiangnan Li, Zheng Lin, Peng Fu, and Weiping Wang.
\newblock Past, present, and future: Conversational emotion recognition through structural modeling of psychological knowledge.
\newblock In Marie{-}Francine Moens, Xuanjing Huang, Lucia Specia, and Scott~Wen{-}tau Yih, editors, {\em Findings of the Association for Computational Linguistics: {EMNLP} 2021, Virtual Event / Punta Cana, Dominican Republic, 16-20 November, 2021}, pages 1204--1214. Association for Computational Linguistics, 2021.

\bibitem{KEC}
Jiangnan Li, Fandong Meng, Zheng Lin, Rui Liu, Peng Fu, Yanan Cao, Weiping Wang, and Jie Zhou.
\newblock Neutral utterances are also causes: Enhancing conversational causal emotion entailment with social commonsense knowledge.
\newblock In Luc~De Raedt, editor, {\em Proceedings of the Thirty-First International Joint Conference on Artificial Intelligence, {IJCAI} 2022, Vienna, Austria, 23-29 July 2022}, pages 4209--4215. ijcai.org, 2022.

\bibitem{ECPEC}
Wei Li, Yang Li, Vlad Pandelea, Mengshi Ge, Luyao Zhu, and Erik Cambria.
\newblock {ECPEC:} emotion-cause pair extraction in conversations.
\newblock {\em {IEEE} Trans. Affect. Comput.}, 14(3):1754--1765, 2023.

\bibitem{DailyDialog}
Yanran Li, Hui Su, Xiaoyu Shen, Wenjie Li, Ziqiang Cao, and Shuzi Niu.
\newblock Dailydialog: {A} manually labelled multi-turn dialogue dataset.
\newblock In Greg Kondrak and Taro Watanabe, editors, {\em Proceedings of the Eighth International Joint Conference on Natural Language Processing, {IJCNLP} 2017, Taipei, Taiwan, November 27 - December 1, 2017 - Volume 1: Long Papers}, pages 986--995. Asian Federation of Natural Language Processing, 2017.

\bibitem{EffectGPT}
Zheng Lian, Licai Sun, Mingyu Xu, Haiyang Sun, Ke~Xu, Zhuofan Wen, Shun Chen, Bin Liu, and Jianhua Tao.
\newblock Explainable multimodal emotion reasoning.
\newblock {\em CoRR}, abs/2306.15401, 2023.

\bibitem{GEVAL}
Yang Liu, Dan Iter, Yichong Xu, Shuohang Wang, Ruochen Xu, and Chenguang Zhu.
\newblock G-eval: {NLG} evaluation using gpt-4 with better human alignment.
\newblock In Houda Bouamor, Juan Pino, and Kalika Bali, editors, {\em Proceedings of the 2023 Conference on Empirical Methods in Natural Language Processing, {EMNLP} 2023, Singapore, December 6-10, 2023}, pages 2511--2522. Association for Computational Linguistics, 2023.

\bibitem{RoBERTa}
Yinhan Liu, Myle Ott, Naman Goyal, Jingfei Du, Mandar Joshi, Danqi Chen, Omer Levy, Mike Lewis, Luke Zettlemoyer, and Veselin Stoyanov.
\newblock Roberta: {A} robustly optimized {BERT} pretraining approach.
\newblock {\em CoRR}, abs/1907.11692, 2019.

\bibitem{ScienceQA}
Pan Lu, Swaroop Mishra, Tanglin Xia, Liang Qiu, Kai{-}Wei Chang, Song{-}Chun Zhu, Oyvind Tafjord, Peter Clark, and Ashwin Kalyan.
\newblock Learn to explain: Multimodal reasoning via thought chains for science question answering.
\newblock In Sanmi Koyejo, S.~Mohamed, A.~Agarwal, Danielle Belgrave, K.~Cho, and A.~Oh, editors, {\em Advances in Neural Information Processing Systems 35: Annual Conference on Neural Information Processing Systems 2022, NeurIPS 2022, New Orleans, LA, USA, November 28 - December 9, 2022}, 2022.

\bibitem{GLUCOSE}
Nasrin Mostafazadeh, Aditya Kalyanpur, Lori Moon, David~W. Buchanan, Lauren Berkowitz, Or~Biran, and Jennifer Chu{-}Carroll.
\newblock {GLUCOSE:} generalized and contextualized story explanations.
\newblock In {\em Proceedings of the 2020 Conference on Empirical Methods in Natural Language Processing, {EMNLP} 2020, Online, November 16-20, 2020}, pages 4569--4586. Association for Computational Linguistics, 2020.

\bibitem{ChatGPT}
OpenAI.
\newblock Chatgpt: Optimizing language models for dialogue.
\newblock 2022.
\newblock Accessed on January 10, 2023.

\bibitem{MELD}
Soujanya Poria, Devamanyu Hazarika, Navonil Majumder, Gautam Naik, Erik Cambria, and Rada Mihalcea.
\newblock {MELD:} {A} multimodal multi-party dataset for emotion recognition in conversations.
\newblock In Anna Korhonen, David~R. Traum, and Llu{\'{\i}}s M{\`{a}}rquez, editors, {\em Proceedings of the 57th Conference of the Association for Computational Linguistics, {ACL} 2019, Florence, Italy, July 28- August 2, 2019, Volume 1: Long Papers}, pages 527--536. Association for Computational Linguistics, 2019.

\bibitem{RECCON}
Soujanya Poria, Navonil Majumder, Devamanyu Hazarika, Deepanway Ghosal, Rishabh Bhardwaj, Samson Yu~Bai Jian, Pengfei Hong, Romila Ghosh, Abhinaba Roy, Niyati Chhaya, Alexander~F. Gelbukh, and Rada Mihalcea.
\newblock Recognizing emotion cause in conversations.
\newblock {\em Cogn. Comput.}, 13(5):1317--1332, 2021.

\bibitem{GPT2}
Alec Radford, Jeffrey Wu, Rewon Child, David Luan, Dario Amodei, Ilya Sutskever, et~al.
\newblock Language models are unsupervised multitask learners.
\newblock {\em OpenAI blog}, 1(8):9, 2019.

\bibitem{SBERT}
Nils Reimers and Iryna Gurevych.
\newblock Sentence-bert: Sentence embeddings using siamese bert-networks.
\newblock In Kentaro Inui, Jing Jiang, Vincent Ng, and Xiaojun Wan, editors, {\em Proceedings of the 2019 Conference on Empirical Methods in Natural Language Processing and the 9th International Joint Conference on Natural Language Processing, {EMNLP-IJCNLP} 2019, Hong Kong, China, November 3-7, 2019}, pages 3980--3990. Association for Computational Linguistics, 2019.

\bibitem{depress}
Sahand Sabour, Wen Zhang, Xiyao Xiao, Yuwei Zhang, Yinhe Zheng, Jiaxin Wen, Jialu Zhao, and Minlie Huang.
\newblock A chatbot for mental health support: exploring the impact of emohaa on reducing mental distress in china.
\newblock {\em Frontiers Digit. Health}, 5, 2023.

\bibitem{TVSHOWGUESS}
Yisi Sang, Xiangyang Mou, Mo~Yu, Shunyu Yao, Jing Li, and Jeffrey~M. Stanton.
\newblock Tvshowguess: Character comprehension in stories as speaker guessing.
\newblock In Marine Carpuat, Marie{-}Catherine de~Marneffe, and Iv{\'{a}}n Vladimir~Meza Ru{\'{\i}}z, editors, {\em Proceedings of the 2022 Conference of the North American Chapter of the Association for Computational Linguistics: Human Language Technologies, {NAACL} 2022, Seattle, WA, United States, July 10-15, 2022}, pages 4267--4287. Association for Computational Linguistics, 2022.

\bibitem{ATOMIC}
Maarten Sap, Ronan~Le Bras, Emily Allaway, Chandra Bhagavatula, Nicholas Lourie, Hannah Rashkin, Brendan Roof, Noah~A. Smith, and Yejin Choi.
\newblock {ATOMIC:} an atlas of machine commonsense for if-then reasoning.
\newblock In {\em The Thirty-Third {AAAI} Conference on Artificial Intelligence, {AAAI} 2019, The Thirty-First Innovative Applications of Artificial Intelligence Conference, {IAAI} 2019, The Ninth {AAAI} Symposium on Educational Advances in Artificial Intelligence, {EAAI} 2019, Honolulu, Hawaii, USA, January 27 - February 1, 2019}, pages 3027--3035. {AAAI} Press, 2019.

\bibitem{SocialIQA}
Maarten Sap, Hannah Rashkin, Derek Chen, Ronan~Le Bras, and Yejin Choi.
\newblock Social iqa: Commonsense reasoning about social interactions.
\newblock In Kentaro Inui, Jing Jiang, Vincent Ng, and Xiaojun Wan, editors, {\em Proceedings of the 2019 Conference on Empirical Methods in Natural Language Processing and the 9th International Joint Conference on Natural Language Processing, {EMNLP-IJCNLP} 2019, Hong Kong, China, November 3-7, 2019}, pages 4462--4472. Association for Computational Linguistics, 2019.

\bibitem{appraisal}
Klaus~R Scherer, Angela Schorr, and Tom Johnstone.
\newblock {\em Appraisal processes in emotion: Theory, methods, research}.
\newblock Oxford University Press, 2001.

\bibitem{CICEROv2}
Siqi Shen, Deepanway Ghosal, Navonil Majumder, Henry Lim, Rada Mihalcea, and Soujanya Poria.
\newblock Multiview contextual commonsense inference: {A} new dataset and task.
\newblock {\em CoRR}, abs/2210.02890, 2022.

\bibitem{ConceptNet}
Robyn Speer, Joshua Chin, and Catherine Havasi.
\newblock Conceptnet 5.5: An open multilingual graph of general knowledge.
\newblock In Satinder Singh and Shaul Markovitch, editors, {\em Proceedings of the Thirty-First {AAAI} Conference on Artificial Intelligence, February 4-9, 2017, San Francisco, California, {USA}}, pages 4444--4451. {AAAI} Press, 2017.

\bibitem{ECE_Survey}
Xinxin Su, Zhen Huang, Yunxiang Zhao, Yifan Chen, Yong Dou, and Hengyue Pan.
\newblock Recent trends in deep learning based textual emotion cause extraction.
\newblock {\em {IEEE} {ACM} Trans. Audio Speech Lang. Process.}, 31:2765--2786, 2023.

\bibitem{LLaMA2}
Hugo Touvron, Louis Martin, Kevin Stone, Peter Albert, Amjad Almahairi, Yasmine Babaei, Nikolay Bashlykov, Soumya Batra, Prajjwal Bhargava, Shruti Bhosale, Dan Bikel, Lukas Blecher, Cristian Canton{-}Ferrer, Moya Chen, Guillem Cucurull, David Esiobu, Jude Fernandes, Jeremy Fu, Wenyin Fu, Brian Fuller, Cynthia Gao, Vedanuj Goswami, Naman Goyal, Anthony Hartshorn, Saghar Hosseini, Rui Hou, Hakan Inan, Marcin Kardas, Viktor Kerkez, Madian Khabsa, Isabel Kloumann, Artem Korenev, Punit~Singh Koura, Marie{-}Anne Lachaux, Thibaut Lavril, Jenya Lee, Diana Liskovich, Yinghai Lu, Yuning Mao, Xavier Martinet, Todor Mihaylov, Pushkar Mishra, Igor Molybog, Yixin Nie, Andrew Poulton, Jeremy Reizenstein, Rashi Rungta, Kalyan Saladi, Alan Schelten, Ruan Silva, Eric~Michael Smith, Ranjan Subramanian, Xiaoqing~Ellen Tan, Binh Tang, Ross Taylor, Adina Williams, Jian~Xiang Kuan, Puxin Xu, Zheng Yan, Iliyan Zarov, Yuchen Zhang, Angela Fan, Melanie Kambadur, Sharan Narang, Aur{\'{e}}lien Rodriguez, Robert Stojnic, Sergey Edunov,
  and Thomas Scialom.
\newblock Llama 2: Open foundation and fine-tuned chat models.
\newblock {\em CoRR}, abs/2307.09288, 2023.

\bibitem{TFD}
Geng Tu, Ran Jing, Bin Liang, Min Yang, Kam{-}Fai Wong, and Ruifeng Xu.
\newblock A training-free debiasing framework with counterfactual reasoning for conversational emotion detection.
\newblock In Houda Bouamor, Juan Pino, and Kalika Bali, editors, {\em Proceedings of the 2023 Conference on Empirical Methods in Natural Language Processing, {EMNLP} 2023, Singapore, December 6-10, 2023}, pages 15639--15650. Association for Computational Linguistics, 2023.

\bibitem{ChatGPTGenKnow}
Geng Tu, Bin Liang, Bing Qin, Kam{-}Fai Wong, and Ruifeng Xu.
\newblock An empirical study on multiple knowledge from chatgpt for emotion recognition in conversations.
\newblock In Houda Bouamor, Juan Pino, and Kalika Bali, editors, {\em Findings of the Association for Computational Linguistics: {EMNLP} 2023, Singapore, December 6-10, 2023}, pages 12160--12173. Association for Computational Linguistics, 2023.

\bibitem{ECF}
Fanfan Wang, Zixiang Ding, Rui Xia, Zhaoyu Li, and Jianfei Yu.
\newblock Multimodal emotion-cause pair extraction in conversations.
\newblock {\em {IEEE} Trans. Affect. Comput.}, 14(3):1832--1844, 2023.

\bibitem{ECTEC_csk}
Fanfan Wang, Jianfei Yu, and Rui Xia.
\newblock Generative emotion cause triplet extraction in conversations with commonsense knowledge.
\newblock In Houda Bouamor, Juan Pino, and Kalika Bali, editors, {\em Findings of the Association for Computational Linguistics: {EMNLP} 2023, Singapore, December 6-10, 2023}, pages 3952--3963. Association for Computational Linguistics, 2023.

\bibitem{summary_LLMs}
Jiaan Wang, Yunlong Liang, Fandong Meng, Beiqi Zou, Zhixu Li, Jianfeng Qu, and Jie Zhou.
\newblock Zero-shot cross-lingual summarization via large language models.
\newblock In Yue Dong, Wen Xiao, Lu~Wang, Fei Liu, and Giuseppe Carenini, editors, {\em Proceedings of the 4th New Frontiers in Summarization Workshop}, pages 12--23, Singapore, December 2023. Association for Computational Linguistics.

\bibitem{ATOMICx10}
Peter West, Chandra Bhagavatula, Jack Hessel, Jena~D. Hwang, Liwei Jiang, Ronan~Le Bras, Ximing Lu, Sean Welleck, and Yejin Choi.
\newblock Symbolic knowledge distillation: from general language models to commonsense models.
\newblock In Marine Carpuat, Marie{-}Catherine de~Marneffe, and Iv{\'{a}}n Vladimir~Meza Ru{\'{\i}}z, editors, {\em Proceedings of the 2022 Conference of the North American Chapter of the Association for Computational Linguistics: Human Language Technologies, {NAACL} 2022, Seattle, WA, United States, July 10-15, 2022}, pages 4602--4625. Association for Computational Linguistics, 2022.

\bibitem{ECPE}
Rui Xia and Zixiang Ding.
\newblock Emotion-cause pair extraction: {A} new task to emotion analysis in texts.
\newblock In Anna Korhonen, David~R. Traum, and Llu{\'{\i}}s M{\`{a}}rquez, editors, {\em Proceedings of the 57th Conference of the Association for Computational Linguistics, {ACL} 2019, Florence, Italy, July 28- August 2, 2019, Volume 1: Long Papers}, pages 1003--1012. Association for Computational Linguistics, 2019.

\bibitem{KI}
Yunhe Xie, Kailai Yang, Chengjie Sun, Bingquan Liu, and Zhenzhou Ji.
\newblock Knowledge-interactive network with sentiment polarity intensity-aware multi-task learning for emotion recognition in conversations.
\newblock In Marie{-}Francine Moens, Xuanjing Huang, Lucia Specia, and Scott~Wen{-}tau Yih, editors, {\em Findings of the Association for Computational Linguistics: {EMNLP} 2021, Virtual Event / Punta Cana, Dominican Republic, 16-20 November, 2021}, pages 2879--2889. Association for Computational Linguistics, 2021.

\bibitem{Bipartite}
Kailai Yang, Tianlin Zhang, Shaoxiong Ji, and Sophia Ananiadou.
\newblock A bipartite graph is all we need for enhancing emotional reasoning with commonsense knowledge.
\newblock In Ingo Frommholz, Frank Hopfgartner, Mark Lee, Michael Oakes, Mounia Lalmas, Min Zhang, and Rodrygo L.~T. Santos, editors, {\em Proceedings of the 32nd {ACM} International Conference on Information and Knowledge Management, {CIKM} 2023, Birmingham, United Kingdom, October 21-25, 2023}, pages 2917--2927. {ACM}, 2023.

\bibitem{Multomodal-CoT}
Zhuosheng Zhang, Aston Zhang, Mu~Li, Hai Zhao, George Karypis, and Alex Smola.
\newblock Multimodal chain-of-thought reasoning in language models.
\newblock {\em CoRR}, abs/2302.00923, 2023.

\bibitem{KBCIN}
Weixiang Zhao, Yanyan Zhao, Zhuojun Li, and Bing Qin.
\newblock Knowledge-bridged causal interaction network for causal emotion entailment.
\newblock In Brian Williams, Yiling Chen, and Jennifer Neville, editors, {\em Thirty-Seventh {AAAI} Conference on Artificial Intelligence, {AAAI} 2023, Thirty-Fifth Conference on Innovative Applications of Artificial Intelligence, {IAAI} 2023, Thirteenth Symposium on Educational Advances in Artificial Intelligence, {EAAI} 2023, Washington, DC, USA, February 7-14, 2023}, pages 14020--14028. {AAAI} Press, 2023.

\bibitem{UECAPrompt}
Xiaopeng Zheng, Zhiyue Liu, Zizhen Zhang, Zhaoyang Wang, and Jiahai Wang.
\newblock Ueca-prompt: Universal prompt for emotion cause analysis.
\newblock In {\em Proceedings of the 29th International Conference on Computational Linguistics, {COLING} 2022, Gyeongju, Republic of Korea, October 12-17, 2022}, pages 7031--7041. International Committee on Computational Linguistics, 2022.

\bibitem{KET}
Peixiang Zhong, Di~Wang, and Chunyan Miao.
\newblock Knowledge-enriched transformer for emotion detection in textual conversations.
\newblock In Kentaro Inui, Jing Jiang, Vincent Ng, and Xiaojun Wan, editors, {\em Proceedings of the 2019 Conference on Empirical Methods in Natural Language Processing and the 9th International Joint Conference on Natural Language Processing, {EMNLP-IJCNLP} 2019, Hong Kong, China, November 3-7, 2019}, pages 165--176. Association for Computational Linguistics, 2019.

\bibitem{TODKAT}
Lixing Zhu, Gabriele Pergola, Lin Gui, Deyu Zhou, and Yulan He.
\newblock Topic-driven and knowledge-aware transformer for dialogue emotion detection.
\newblock In Chengqing Zong, Fei Xia, Wenjie Li, and Roberto Navigli, editors, {\em Proceedings of the 59th Annual Meeting of the Association for Computational Linguistics and the 11th International Joint Conference on Natural Language Processing, {ACL/IJCNLP} 2021, (Volume 1: Long Papers), Virtual Event, August 1-6, 2021}, pages 1571--1582. Association for Computational Linguistics, 2021.

\end{thebibliography}
\end{document}